\documentclass{article}

\usepackage{arxiv}
\usepackage{setspace}

\usepackage{times}
\usepackage{latexsym}
\usepackage{tabularray}
\usepackage{tabularx}
\usepackage[T1]{fontenc}
\usepackage[utf8]{inputenc}

\usepackage{microtype}

\usepackage{inconsolata}

\usepackage{graphicx}
\usepackage{placeins}

\usepackage{amsmath}

\usepackage{subfigure}
\usepackage{booktabs}
\usepackage{graphicx}
\usepackage{subcaption}
\usepackage{fvextra}
\usepackage[most]{tcolorbox}

\newtcolorbox{promptbox}{
    colback=gray!5,
    colframe=gray!50,
    boxrule=0.8pt,
    arc=2pt,
    left=6pt,
    right=6pt,
    top=4pt,
    bottom=4pt,
    before upper={\singlespacing}
}

\usepackage[
  backend=biber,
  style=apa,      % gives you author-year like apalike
  natbib=true,           % keeps \citep, \citet, \citealt etc. working
  maxbibnames=99,        % list all authors in the bibliography (biblatex defaults to truncating)
  maxcitenames=2,        % "et al." after 2 names in citations—adjust to taste
  uniquename=false
]{biblatex}
\DeclareCaseLangs{}

\usepackage{caption}
\usepackage{url}
\usepackage{hyperref}
\usepackage{makecell}

\title{When Persona Attributes Improve Population Alignment in Large Language Models}
\date{}

\newcommand\blfootnote[1]{%
  \begingroup
  \renewcommand\thefootnote{}\footnote{#1}%
  \addtocounter{footnote}{-1}%
  \endgroup
}

\usepackage{authblk}

\author[1,*]{%
	Leon Fröhling
}
\author[2,*]{%
	Jens Rupprecht
}
\author[1,2,3]{%
	Markus Strohmaier
}
\author[1,3,4]{%
	Claudia Wagner
}
\affil[1]{GESIS – Leibniz Institute for the Social Sciences}
\affil[2]{University of Mannheim}
\affil[3]{Complexity Science Hub}
\affil[4]{RWTH Aachen University}

\begin{document}

\maketitle

\begin{abstract}
    Large Language Models (LLMs) are increasingly used to predict the responses of human participants in survey panels. Towards that goal, persona prompting has recently emerged as a technique to inform and align large pretrained language models. Persona prompting refers to the practice of using short textual descriptions of 'personas' in prompts to steer the LLM's generations. Personas describe individuals through different attributes such as their socio-demographics, attitudes, or behaviors, with the aim of aligning LLMs to produce responses that correlate with the corresponding human responses.
    Yet, recent work has produced mixed and partly conflicting results of persona prompting without clear patterns of success and failure. Among the few consistent findings is that the selection of persona attributes matters, and that using more attributes does not necessarily lead to better performance. It remains unclear how different attribute selection methods perform and how to choose among them.
    In this paper, we propose that observed human response variation of a survey question is a potential explanation for the mixed performance observed so far. In addition, we compare the performance of persona prompting associated with different methods for selecting persona attributes. We evaluate these methods on four different (general) social surveys across two countries, six LLMs, and twenty prediction tasks per survey. Our work helps to identify \textit{when} persona prompting can be expected to be useful in survey prediction tasks, and provides new insights on the effectiveness of different attribute selection methods for LLM-based survey prediction using persona prompting.
\end{abstract}
\blfootnote{*~Equal contribution.}

\section{Introduction}

\begin{figure}
    \centering
    \includegraphics[alt={A graphical depiction of our methodology and experimental setup.}, width=0.9\linewidth]{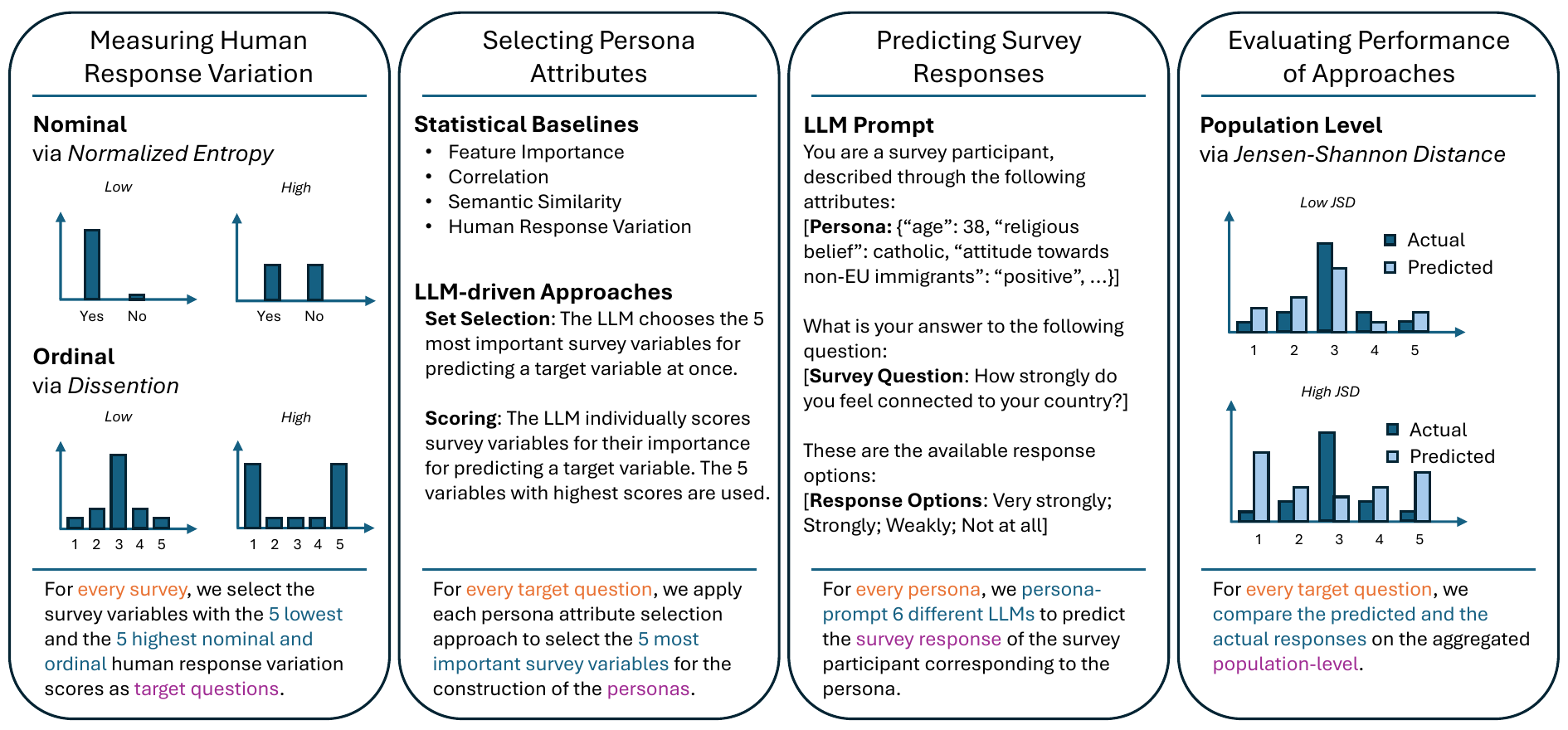}
    \caption{\textbf{Overview and Experimental Setup.} Using survey data, we evaluate how well different persona attribute selection strategies help to predict responses of survey participants to questions with differing levels of human response variation. We hypothesize that persona-prompted LLMs are better at predicting responses to survey question with higher human response variation, and that persona attribute selection approaches optimized for persona prompting outperform purely data-driven alternatives.}
    \label{fig:overview}
\end{figure}

The use of personas to introduce diversity and simulate different human perspectives is rapidly gaining traction in Natural Language Processing (NLP) and Computational Social Science (CSS). By conditioning large language models (LLMs) on demographic, attitudinal, or behavioral characteristics, persona prompting promises a scalable approach to modeling heterogeneous viewpoints and social behavior, often combining empirical observations with synthetic data generation.

Despite this promise, empirical findings remain mixed. While some studies report substantial improvements in the representation of diverse perspectives, others find limited or inconsistent gains, raising fundamental questions about when and why persona prompting succeeds. In particular, the conditions under which personas enable LLMs to faithfully reproduce human judgments and behaviors remain poorly understood.

The limited empirical evidence available on the sources of these mixed results comes primarily from subjective annotation tasks. In this context, \citet{brown2025evaluating} and \citet{oluyemi2024corpus} show that the degree of agreement among human annotators predicts an LLM's ability to reproduce human annotations. Their findings suggest that persona prompting is more effective when human judgments are relatively consistent, but less successful when opinions diverge substantially. While these studies provide important initial insights, it remains unclear whether this relationship generalizes beyond the specific annotation tasks that they analyzed.

To study this question in the context of surveys, we introduce the concept of human response variation, which we define as the variation observed in participants' responses to a given survey question. This conceptualization is closely related to the notion of human label variation in NLP \citep{plank2022problem}, which recognizes that variation in annotations may arise not only from task ambiguity but also from legitimate differences in interpretation and multiple plausible responses. Survey responses, however, differ in an important respect: variation is typically not a measurement artifact but the substantive phenomenon of interest. Respondents differ in their socio-demographic characteristics, opinions, attitudes, experiences, and behaviors, all of which can give rise to systematically different yet equally valid responses to the same question. Consequently, our concept of human response variation captures meaningful heterogeneity in the population. We propose differences in human response variation as a potential explanation for performance differences of the persona prompting approach in reproducing human response distributions across different survey response prediction tasks.

In addition to being a promising application, surveys provide the ideal evaluation environment for methodological advances: They provide detailed individual-level information ranging from basic socio-demographic characteristics to more complex behaviors and beliefs from which relevant attributes for the construction of persona descriptions can be selected, and naturally feature a range of diverse questions with varying levels of human response variation that can be used for evaluation. We operationalize human response variation through two different measures of dispersion, selected based on the type and scale of the response variable \citep{wilcox2008variance}: normalized entropy for nominal and dissention for ordinal response variables.

We build upon one of the few consistent findings in the literature on persona prompting to further understand the factors influencing the performance of the approach: Selecting the right set of attributes to construct persona descriptions matters. While \citet{luz2025principled} and \citet{rupprecht2025german} observe that LLMs are sensitive to irrelevant persona details, with performance dropping sharply if they are included, \citet{hwang2023aligning} find that augmenting a set of demographic attributes and ideology scores with past user opinions leads to consistent accuracy gains across a range of tasks. Notably, previous work either added or removed whole blocks of attributes irrespective of the predictive task at hand. Instead, we benchmark different approaches for selecting persona attributes given a target survey question to predict responses for. Throughout this work, we refer to the prediction of responses for specific survey questions as survey response prediction tasks, given that both the prediction target and the inputs vary. 

In this work, we address the following two research questions (RQs):
\begin{itemize}
    \item \textbf{RQ1}: To what extent can human response variation explain the differences in persona prompting performance across different survey response prediction tasks? 
    \item \textbf{RQ2}: To what extent can different persona attribute selection methods improve the performance of persona prompting on survey response prediction tasks?
\end{itemize}

To address these research questions, we systematically investigate the relationship between human response variation, persona variable selection methods, and persona prompting performance, using the methodology outlined in Figure~\ref{fig:overview}. We use national-level data for the US and Germany from an international survey program run in parallel in 70 countries across the world, the World Value Survey (WVS) \citep{evswvs2024}, as well as the general social surveys conducted in these two countries \textemdash the US General Social Survey (GSS) \citep{gss2025}, and the German General Social Survey (GGSS) \citep{ggss2025}.

\section{Background}
\label{sec:background}

\subsection{Persona Prompting Studies Report Mixed Results}

Persona prompting has widely been used to simulate human responses to survey questions. However, the findings remained mixed: for example, some work found that they were not able to recover known regression coefficients by conditioning LLMs with persona prompting \citep{bisbee_synthetic_2024}, while others state the ambiguity of results \citep{lee_can_2024,qu_performance_2024,sanders_demonstrations_2023}. At the same time, there is a large amount of work achieving promising results with persona-conditioned LLMs on various tasks \citep{park2024generative,aher_using_2023,argyle_out_2023,horton_large_2023, li_persona-based_2026}. 

Even though there already exists a wide range of applications for personas and methods for constructing them, only few previous works have set out to understand the reasons for why or why not the approach works in a given context. In the context of NLP, some authors have started to explore the connection between the selection of persona variables, the variation in human annotations, and the performance of persona prompting. \citet{hu2024quantifying} fit linear regressions to predict annotations using persona variables, finding that those variables usually explain less than 10\% of variance across NLP tasks. When using these persona variables for persona prompting to replicate the tasks’ annotations, they find that variables that are most strongly correlated with the outcome measure lead to the best predictive performance, and that most gains are to be expected for tasks with high entropy but low standard deviation in human annotations. Furthermore, \citet{hu2025simbench} establish a link between the stage of model development and the models’ ability to represent differing levels of variation human behavior. They distinguish between base models, optimized for "[representing] the full, multi-modal diversity of human language and opinion during pre-training", and instruction-tuned models, optimized for "[concentrating] probability mass on a single, high-reward mode of the target preference distribution". They find that instruction-tuned LLMs working best for tasks with low-entropy outcomes and base models for tasks with high-entropy outcomes, suggesting that instruction-tuned models are better at accurately representing majority outcomes while base models are better able to replicate the full distribution of human behavior.

Taking a different angle, \citet{hwang2023aligning} use the response variation of human participants with the same basic demographic attributes to determine the degree to which additional attributes are needed to explain the observed response variation. They show that the agreement scores (measured via Cohen’s $\kappa$) between individuals with the same demographics are gathered around 0.5, with perfect agreement being equal to scores of 1. This finding indicates that using only this set of demographics is insufficient for modeling the disagreement, and that therefore additional attributes are needed for meaningfully explaining the differences in outcomes.

In addition to this empirical evidence from NLP research, suggesting that human response variation is a plausible explanation for the mixed performance of persona prompted LLMs in the survey response prediction task, we derive two complementary explanations from general observations on the development and use of LLMs.

Similar to the observations of \citet{hwang2023aligning}, we hypothesize that persona prompting works best for predicting survey questions with high levels of human response variation, because the LLM requires the additional attributes available through the persona description to model deviations from its default response, i.e., the response it would generate without being conditioned on a persona. Intuitively, if two individuals with different attributes respond differently to a survey question, an LLM used for predicting their individual survey responses would need access to these differing attributes to be able to meaningfully model the differences in their responses.

Similar to the observations of \citet{hu2025simbench}, we hypothesize that persona prompting works worst for predicting survey questions with low levels of human response variation, because the optimal response generation strategy for the LLM would be to simply generate its default response. As has been shown empirically and derived theoretically, the default response of LLMs tends towards a representation of the majority view \citep{chakraborty2024maxmin, santurkar2023whose, sorensen2024position}. In this view, any additional persona attribute introduces noise and has the potential to steer the LLM away from its majority-aligned default response, thus decreasing performance in cases in which the response distribution is highly concentrated around a single majority response. \citet{hu2025simbench} observe that this effect is particularly strong for instruction-tuned models, which they describe as optimized for producing this type of single best response instead of a full distribution of (equally) valid responses.

With respect to our first research question, we thus hypothesize that the human response variation of the survey question to be predicted has an impact on the performance of the persona prompting approach. We expect the approach to work better for questions with high human response variation versus for questions with low human response variation, and that this effect is reversed when not using personas.

\subsection{Persona Attribute Selection Matters}

In one of the few methodological contributions on the mechanism of persona prompting, moving beyond external factors potentially explaining performance differences, \citet{luz2025principled} point to the gap in the literature that we are aiming to address: "Despite a diversity of personas and tasks, most prior work does not systematically differentiate between relevant and irrelevant persona attributes or measure their specific influence on model behavior".

In their own work, \citet{luz2025principled} formulate four desiderata of persona prompting, of which \textit{task-irrelevant attributes should not affect model performance} and \textit{relevant attributes [..] should shape model performance in ways consistent with those attributes} stand out as particularly important for our context. For a number of objective annotation tasks, they find that personas differentiated by relevant expertise and education mostly impact task performance in expected ways, and that the inclusion of irrelevant attributes such as color preferences and names leads to drops in performance of up to 30 percentage points.

\citet{giorgi2024modeling} study how explicit (defined via their age, gender, political ideology, race, and substance use) and implicit (defined via names that are highly frequent for specific demographics) personas in LLM prompts replicate patterns of human subjectivity in annotation and generation tasks. They find that explicitly providing attributes rather than having the LLM infer them from names works better, and that political ideology seems to be the driving factor behind the persona-conditioned generations, with gender being the least important one. Similar to the distinction between explicit and implicit personas by \citet{giorgi2024modeling}, there are a number of works studying the impact of different formats or locations for presenting the persona information in the prompt, with few consistent findings across these different studies \citep{lutz2025prompt, neumann2025position, weeber2026one}.

In the survey context, \citet{xie2026evaluating} generate responses to held-out survey questions in order to evaluate whether LLMs conditioned on demographic variables are successful in generating digital twins of real populations. They observe that including more attributes in the conditioning prompts increases statistic realism, i.e., improves the ability of persona-prompted LLMs to reproduce the actual human responses. Similarly, \citet{park2024generative} have shown that conditioning LLM-generations on rich qualitative data such as semi-structured interviews leads to the best performance in predicting participants’ survey responses. However, they also find that using structured survey data for persona prompting works similarly well. In contrast, the performance drops significantly when using only the set of four commonly-used socio-demographic attributes age, gender, race and political ideology for persona construction.

\citet{hwang2023aligning} use embedding similarity between input attributes (in their case individuals’ past opinions) and the target question to construct personas from sets of attributes with differing importance, assuming that higher similarity between attribute and target question implies higher attribute importance. They find that using the top-3 attributes performs on par with using the combination of socio-demographics, ideology, and 16 randomly selected attributes, suggesting that selecting the most important attributes for persona prompting matters.

In contrast to previous work on persona prompting that either relies on a small set of socio-demographic attributes, augmented with manually selected, task-dependent additions such as political ideology or past opinions \citep{giorgi2024modeling, hwang2023aligning}, or that selects the same set of attributes across different tasks \citep{rupprecht2025german}, we benchmark different approaches for selecting an optimal set of attributes for each survey response prediction task.

We are particularly interested in comparing the performance of LLM-based approaches with that of statistical baseline approaches. We design the statistical baseline approaches to select persona attributes exclusively based on correlations and associations directly available from the full survey data. In contrast, for the LLM-based approaches, we hope to leverage the broad world knowledge that these models develop during the different stages of pre-training and alignment and that enables them to generalize to novel contexts and tasks, including survey response prediction \citep{wang2024generalization, chen2023skills}. However, because LLMs’ internal knowledge representations are different from those of humans, the relationships and correlations leveraged by LLMs during response prediction are not necessarily evident from an isolated survey and cannot fully be captured by traditional statistical techniques \citep{lappin2024assessing, suresh2023conceptual}.

By explicitly prompting for a selection of optimal sets of attributes for use in a persona prompting application, we hope that the LLM-based attribute selection approaches leverage their internal representations of knowledge to select LLM-optimized attributes. With respect to our second research question, we assume that these LLM-optimized sets of selected attributes should lead to better performance of the persona prompting approach than using attributes selected by statistical baselines that rely on survey data alone.

\section{Methodology}
\label{sec:methodology}

In this section, we present the different parts of our methodological setup, represented by the panels in Figure~\ref{fig:overview}. We start by introducing our measures of human response variation as well as the persona attribute selection approaches we are testing. We describe how personas are constructed and used to predict human survey responses and conclude with our measures to evaluate and compare the performance of the different approaches.

\subsection{Measuring Human Response Variation}
\label{sec:methodology_hrv}

As discussed in the introduction, we propose to use the human response variation observed in the human responses to a survey question as a measure for the difficulty that a persona-prompted LLM is facing in correctly predicting the human responses to the given question. To be able to compare the human response variation across different survey variables~\footnote{We are generally using \textit{survey variable} to refer to the combination of survey question and the (human) responses to it,  and \textit{survey question} to refer to the text of the question and the device to solicit responses.}, we require a measure that summarizes the spread of human responses across varying number of response options. 

For survey variables with \textbf{nominal responses}, such as a binary yes-or-no response or a preference between different political parties, we can simply use the \emph{normalized Shannon entropy} \citep{shannon1948mathematical} as a measure for the average uncertainty of the variable's potential outcomes.

The entropy of a (survey response) distribution is given as $$H(X)=-\sum_{i=1}^{n}p(X_i)\log_2 p(X_i)\text{,}$$ with $X_i\in[X_1,...,X_n]$ as the set of all possible response options and $p(X_i)$ as the relative frequency of each response. The normalized entropy is then calculated by dividing the entropy by the maximal possible entropy associated with the response variable, which is given as $H_{max}=\log_2 n$.

However, survey variables with \textbf{ordinal responses}, such as Likert scales with responses ranging from, e.g., strongly disagree to strongly agree, cannot be analyzed by the Shannon entropy as it is invariant to the location of responses on the scale. Thus, \citet{tastle2006information} develop their consensus measure, which, if inverted, becomes the \emph{dissention measure}, measuring the human response variation in ordinal variables. 

The only requirement of this measure is the ability to identify the extreme values associated with the endpoints of the range of possible responses. Once these are defined, a numerical scale is applied to the response options, assigning a value of $1$ to one endpoint and a value equivalent to the number of response options to the other endpoint. All response options in between are assigned integer values increasing in unit steps. In the case of a five-point Likert scale ranging from strongly disagree to strongly agree, strongly disagree would be equal to $1$, disagree to $2$, neutral to $3$, agree to $4$, and strongly agree to $5$. 

This first transformation results in $X=\{1,2,3,4,5\}$, with $X_1=1$, $X_2=2$, and so on. The width of $X$ is given as $d_X=X_{max}-X_{min}$, and the expected value as $$E_X=\sum_{i=1}^{n}p(X_i)X_i=\mu_X\text{,}$$ with $n$ as the number of response options (i.e., the length of $X$) and $p(X_i)$ the relative frequency associated with each response option $X_i$.

Finally, the consensus measure is calculated as $$\text{Cns}(X)=1+\sum_{i=1}^{n}p(X_i)\log_2(1-\frac{|X_i-\mu_X|}{d_X})\text{.}$$ The inverse gives the measure of dissention, $\text{Dis}(X)=1-\text{Cns}(X)$. 

Both the normalized Shannon entropy for nominal and the dissention for ordinal variables have the same crucial properties: They are bound between $0$ and $1$, with values towards $0$ corresponding to minimum human response variation and values towards $1$ corresponding to maximum human response variation. While both measures take values in the same range, they are not directly comparable. We therefore select \textit{target variables} \textemdash the set of survey variables we hold out to predict human responses for \textemdash based on the normalized entropy and dissention scores for nominal and ordinal survey variables separately.

\subsection{Selecting Persona Attributes}
\label{sec:methodology_personas}

\begin{table}[]
    \renewcommand{\arraystretch}{1.5} % increase spacing between table-rows
    \centering
    \begin{tabular}{p{2cm}p{12cm}}
        \hline
        \multicolumn{2}{l}{Baseline Approaches} \\
        \hline
        Feature Importance & We use the same approach as \citet{rupprecht2025german}, extracting feature importance scores from random forest classifiers fitted for the target variable, using all remaining variables as input features. We select the five most important features as the set of selected attributes.\\
        Correlation & For the target variable, we calculate correlations with every remaining variable, selecting the five variables with the highest correlations to the target variable as the set of selected attributes.\\
        Semantic Similarity & We identify those survey questions that are semantically most similar to the target question as potentially most relevant for predicting it. We calculate semantic similarity based on embeddings of the survey questions produced using the \textsc{all-MiniLM-L6-v2} available via \textsc{SentenceTransformers}~\footnote{\url{https://sbert.net/docs/sentence_transformer/usage/usage.html}}, taking the five survey variables with question texts closest to that of the target variable as the set of selected attributes.\\
        Human Response Variation & We select the five variables with the highest human response variation scores as the set of selected attributes, following the traditional use of entropy as a measure of information \citep{shannon1948mathematical}. Note that the human response variation of survey variables is independent of a given target variable\textemdash therefore, and in contrast to all other approaches, this approach results in a single set of attributes that is used across all target variables.\\
        \hline
        \multicolumn{2}{l}{LLM-based Approaches} \\
        \hline
        Set Selection & The LLM is given the target question text as well as a list of all other variables' question texts from the same survey. The LLM is then prompted to select the five most important survey variables for predicting individual-level survey responses to the target variable.\\
        Scoring & The LLM is given the text of the target question as well as the text of a single question from the same survey. The LLM is then prompted to score the importance of the corresponding survey variable for predicting individual-level survey responses to the target variable on a range from 0 ("not important at all") to 100 ("most important"). This is repeated for all variables in the survey, resulting in a target-variable-specific importance ranking of survey variables. Based on this ranking, the five top-ranked variables are selected for persona construction.\\
    \end{tabular}
    \caption{\textbf{Persona attribute selection methods.} Baseline and LLM-based approaches for selecting persona attributes for a given target variable. Apart from the Semantic Similarity approach, all baseline approaches have the full survey data as input. In contrast, the LLM-based approaches only have access to the question texts of the target variable as well as those of the remaining survey variables.}
    \label{tab:selection_approaches}
\end{table}

In this subsection, we describe the different approaches we are benchmarking for the selection of an optimal set of persona attributes for each target variable. In contrast to, e.g., \citet{rupprecht2025german}, we do not aim to produce a single set of attributes that is supposed to be optimal across all selected target variables, but we design our approaches to identify an optimal set of attributes for each target variable independently.

For \textbf{RQ2}, our main focus is on the comparison of statistical baseline approaches and LLM-based approaches for the selection of persona attributes. While the baseline approaches rely on the survey data and are agnostic towards the use of LLMs as the system for predicting survey responses, the LLM-based approaches are explicitly prompted to select the optimal set of attributes for predicting individual-level survey responses using a zero-shot persona-prompted LLM. Table~\ref{tab:selection_approaches} further describes the different selection approaches, with the prompts used for the LLM-based approaches being documented in Appendix~\ref{app:prompts_selection}.

Naturally, we prevent the selection of survey variables that are variants of the target variable as persona attributes by removing all duplicate survey variables during pre-processing of the survey data (see Section~\ref{sec:experiments_data}) and by excluding the target variables from the list of available survey variables during persona attribute selection.

\subsubsection{Stability} 
\label{sec:methodology_stability}

To check for the stability of the (non-deterministic) LLM-based selection approaches, we run them 100 times for each constellation of target question and LLM. For the two largest models, Qwen2.5-VL-32B and Llama3.3-70B, we reduce the number of runs to 30 due to computational resource constraints. Between runs, we shuffle the order in which the survey variables are presented to the LLM. Particularly for the set selection approaches, which fit a large number of survey variables into a single prompt, the LLM might have difficulties to put equal attention on all positions of the context window. Therefore, we expect the selected sets to differ between runs. 

For the construction of personas based on the multiple runs of the LLM-based approaches, we use the five survey variables featured most often in the selected sets across runs for the set selection approach, and the five survey variables with the highest average score across runs for the scoring approach. We report our approaches for evaluating the stability of the LLM-based attribute selection approaches in Appendix~\ref{app:results_stability}.

\subsection{Predicting Survey Responses}
\label{sec:methodology_prediction}

To not interfere with our evaluation of the different persona attribute selection approaches, we do not vary the prompts and generation parameters used for the actual survey response prediction. We also focus on zero-shot persona prompting, leaving the evaluation of alternative approaches such as few-shot prompting and fine-tuning for future work. 

To predict the survey response of a survey participant for a given target variable, we insert the persona description constructed from the set of selected persona attributes, the text of the target question, as well as the range of possible response options as input into a structured prompt, asking the persona-prompted LLM to generate the most likely response given the persona description as output. 

As the no-persona baseline for answering \textbf{RQ1}, we repeatedly prompt the LLM to predict responses to the given target variable without specifying a particular persona. We repeat this as many times as we have participants in a given survey. We provide more details on the response generation process as well as the prompts we use in Appendix~\ref{app:prompts_prediction}.

\subsection{Evaluating Survey Response Prediction Performance}
\label{sec:methodology_evaluation}

We evaluate the performance of the persona selection approaches on the aggregated population-level. We compare the response distribution resulting from aggregating the individual-level response predictions of all personas with the human response distribution resulting from aggregating the responses of all survey participants for a given target question. 

We use the Jensen-Shannon Distance (JSD) to assess how well the predicted responses approximate the human responses across the entire set of participants in a survey. The Jensen-Shannon Distance is normalized to the range from $0$ to $1$. Lower values indicate higher similarity between the two distributions and thus better performance of the approach in predicting the response distribution.

\section{Experimental Setup} 
\label{sec:experiments}

As laid out above, surveys offer the ideal environment for developing persona collections and evaluating the underlying methodology. We describe the different surveys we use in this work as well as the steps necessary to prepare them for the use in our persona prompting setup. We report the target variables we select for evaluation based on the measures of human response variation and discuss our choice of models for predicting survey responses.  

\subsection{Survey Data}
\label{sec:experiments_data}

\begin{table}
    \centering
    \begin{tabular}{lc|rr|rrrrrrr|rr}
        Name    & Year  & $N$   & $Q$       & $Q^{T}_{SD}$  & $Q^{T}_{BC}$  & $Q^{T}_{AT}$  & $Q^{T}_{RE}$  & $Q^{T}_{CO}$  & $Q^{T}_{NP}$  & $Q^{T}_{D}$  & $Q^{S}_{N}$   & $Q^{S}_{O}$   \\  
        \hline
        GGSS    & 2023  & 5,246 & 582       & 52            & 95            & 224           & 157           & 3             & 51            & 0            & 66            & 247           \\
        GSS     & 2024  & 3,309 & 207       & 38            & 31            & 74            & 7             & 4             & 31            & 22            & 51            & 54            \\
        WVS-DE  & 2018  & 1,528 & 348       & 31            & 60            & 203           & 23            & 3             & 25            & 3            & 64            & 199           \\
        WVS-US  & 2017  & 2,596 & 337       & 28            & 57            & 203           & 16            & 3             & 27            & 3            & 64            & 196           \\
        \\
    \end{tabular}
    \caption{\textbf{Survey datasets used in our study.} We use the latest available data from four surveys collected in two different countries (GGSS and WVS-DE from Germany, GSS and WVS-US from the US). $N$ denotes the number of survey participants and $Q$ the number of variables in each of the four surveys. $Q^{T}_{SD}$, $Q^{T}_{BC}$, $Q^{T}_{AT}$, $Q^{T}_{RE}$, $Q^{T}_{CO}$, and $Q^{T}_{NP}$ are the number of variables annotated as the different types: socio-demographic (SD), behavioral/circumstantial (BC), attitudinal (AT), relational (RE), contextual (CO), non-participant (NP), and duplicate (D). $Q^{S}_{N}$ and $Q^{S}_{O}$ are the number of variables of types BC and AT annotated as being nominal (N) or ordinal (O), i.e., the survey variable from which we select target variables for evaluation. The discrepancy between $Q^{T}_{BC}+Q^{T}_{AT}$ and $Q^{S}_{N}+Q^{S}_{O}$ for GGSS is due to the presence of six continuous variables, which we do not consider for the selection of target variables.}
    \label{tab:surveys}
\end{table}

We run the same set of experiments on data from different surveys in order to ensure the robustness and generalizability of our results. Additionally, we study two different socio-cultural contexts by using surveys from the United States and Germany. For both countries, we work with the respective general social survey \textemdash the General Social Survey (GSS) for the United States, and the German General Social Survey (GGSS) for Germany. By also using the data from the World Value Survey (WVS) for these two countries, we include an additional survey that is directly comparable across the different countries it is run in. We document the steps necessary for preprocessing the survey data as well as the annotations of variable type and scale in Appendix~\ref{app:experimental_setup_data}. We report the number of variables featured in the different surveys as well as the results of our annotation efforts in Table~\ref{tab:surveys}.

\subsection{Target Questions}
\label{sec:experiments_targets}

\begin{figure}
    \subfigure[Distribution of Human Response Variation as measured via normalized entropy for \textbf{nominal} response variables.]{\includegraphics[alt={Histograms of the human response variation for the nominal variables in four surveys.}, width=\linewidth]{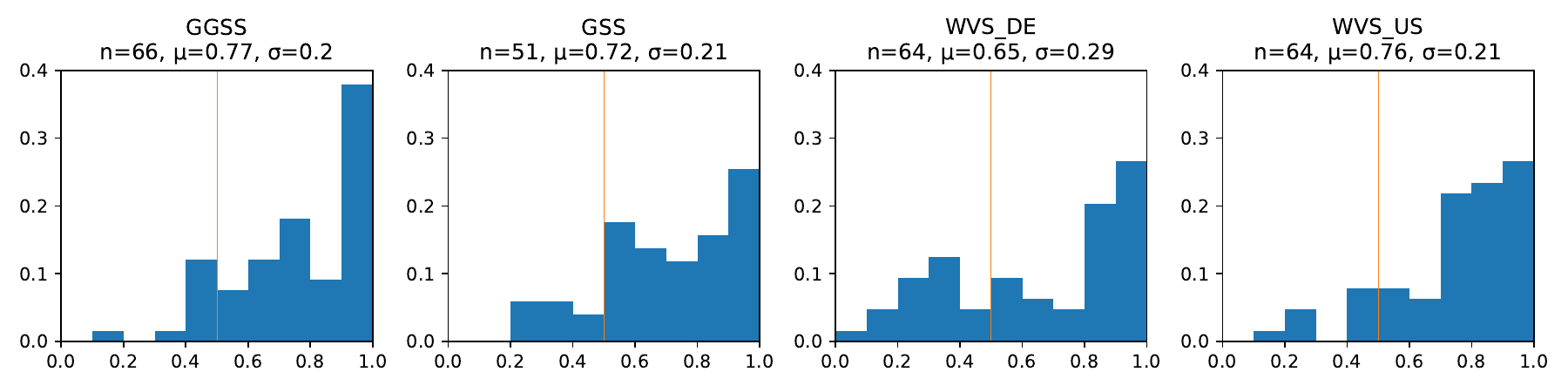}}
    \subfigure[Distribution of Human Response Variation as measured via dissention scores for \textbf{ordinal} response variables.]{\includegraphics[alt={Histograms of the human response variation for the ordinal variables in four surveys.}, width=\linewidth]{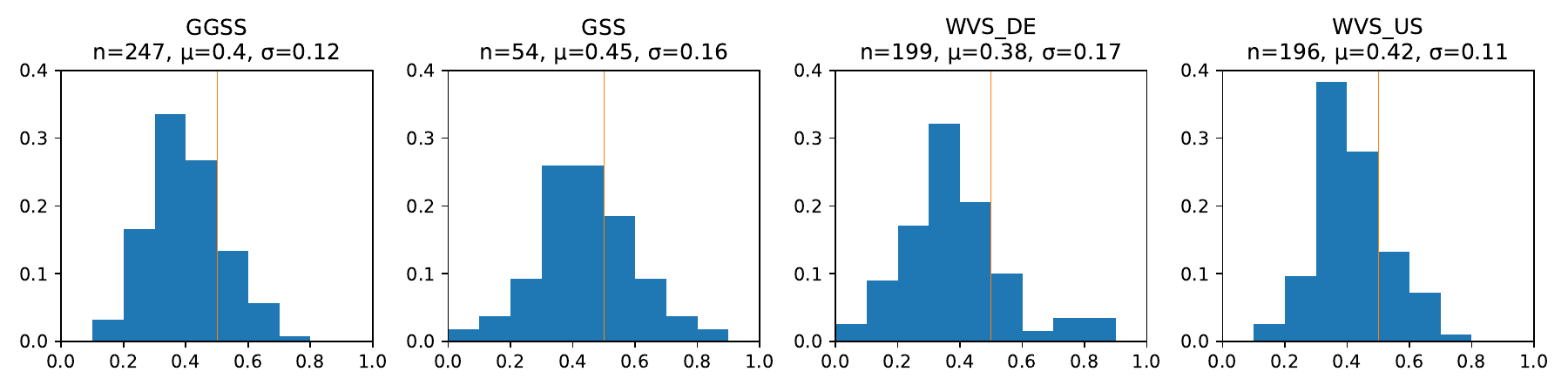}}
	\caption{\textbf{Distribution of human response variation across surveys for nominal (a) and ordinal (b) variables.} Only variables of type behavioral/circumstantial and attitudinal are included, as we consider only those types of variables as potential target variables. The x-axis shows the human response variation, with higher values corresponding to higher levels of variation in the human responses to a given survey question. The y-axis shows the relative frequency of the buckets of widths 0.1. For each survey, we select the five nominal and the five ordinal variables with lowest and highest human response variation scores as target questions.}
	\label{fig:disagreement}
\end{figure}

To study the effects of the human response variation levels of different survey variables on the ability of persona-prompted LLMs to correctly predict survey responses, we systematically select a set of target variables for evaluation of our experiments.

First, we consider only survey variables which were annotated as representing behavioral/ circumstantial or attitudinal information \textemdash these types of variables represent the much more interesting use case in the context of social science and survey research, compared to the prediction of socio-demographic (e.g., the individual's age), relational (e.g., the profession of the individual's father), contextual (e.g., the year in which the survey was taken), or non-participant (e.g., the interviewer's age) survey responses.

Next, we calculate the appropriate measure of human response variation given the scale of the survey variable. In doing so, we produce a separate ranking of human response variation scores for nominal and ordinal variables. For both nominal and ordinal variables, we select the five variables with lowest and the five variables with highest human response variation scores as target variables. We exclude survey variables which have been answered by less than 500 participants from this selection to ensure sufficient data for evaluation.

Figure~\ref{fig:disagreement} shows the distributions of human response variation scores across the four different surveys. Appendix~\ref{app:experimental_setup_targets} Tables~\ref{tab:target_questions_gss} to \ref{tab:target_questions_wvs_us} document the target variables that were selected based on their low and high human response variation scores, respectively. Appendix~\ref{app:experimental_setup_targets} Figures~\ref{fig:targets_nominal_low_HRV} to \ref{fig:targets_ordinal_high_HRV} show the response distributions of these target variables.

\subsection{Model Selection}
\label{sec:experiments_models}

In the base configuration of our experiments, we always use the same LLM for the two steps potentially involving LLMs: the selection of persona attributes and the prediction of survey responses. We choose models from different model families and providers (Qwen models from Alibaba Cloud and Llama models from Meta AI) and with different parameter sizes (from 3B to 70B) in order to investigate the impact of model choice on the performance and to ensure generalizability of the persona prompting approach for survey response prediction. All models we use are available via Huggingface~\footnote{\url{https://huggingface.co/}} and we use the vLLM~\footnote{\url{https://docs.vllm.ai/en/stable/}} framework for interactions with models. Appendix~\ref{app:experimental_setup_models} Table~\ref{tab:model_releases} gives an overview of the different models used as well as their (known) training cutoff dates. Data leakage from the surveys into the models we are using is practically impossible for GSS and GGSS and highly unlikely for the two WVS survey datasets.

\section{Results}
\label{sec:results}

We present the experimental results of applying the complete methodological pipeline shown in Figure~\ref{fig:overview} to the four selected surveys. While we generally observe that larger LLMs tend to perform better across attribute selection approaches, we focus our discussion of the results on the two research questions, separating them into disaggregated views for the different model families and surveys.

\begin{figure}
	\centering
    \includegraphics[alt={Data points and mean lines that compare the performance of different LLMs that were prompted with persona descriptions constructed through different attribute selection methods and without persona descriptions.}, width=\linewidth]{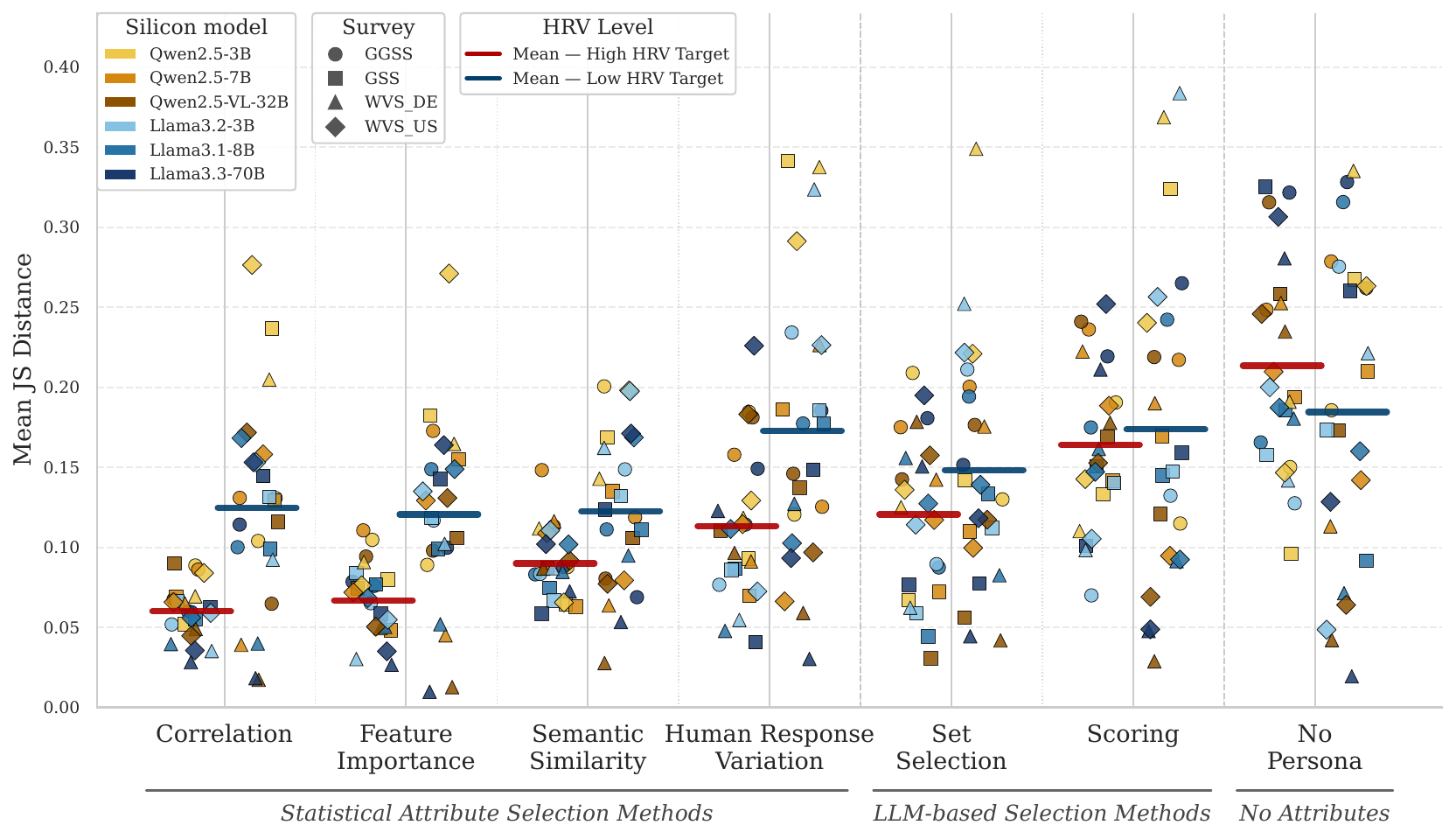}
	\caption{\textbf{Overview of our results. Persona prompting works better for questions with high human response variation and statistical baselines outperform LLM-based approaches for attribute selection.} Each data point in this figure represents the averaged performance of a persona-prompted LLM in predicting the responses to 20 target questions (differentiated by position of data points and mean line, with high HRV questions oriented to the left and with red mean line, and low HRV questions oriented to the right and with blue mean line) from one of four different surveys (differentiated by shape, see legend). Along the x-axis are the different approaches used for selecting the persona attributes for persona prompting as well as the LLM baseline without persona conditioning (\textit{No Persona}). The y-axis indicates the performance measured by the Jensen-Shannon Distance (lower is better) between the predicted and the actual response distributions across all participants in a given survey. The main takeaways from our experiments are that the persona prompting setups are better for predicting responses for target questions with high human response variation across all attribute selection approaches [red mean lines consistently lower than blue mean lines; \textbf{RQ1}], and that using statistical baselines based on the survey data alone (\textit{Correlation} and \textit{Feature Importance}) for selecting persona attributes lead to better persona prompting performance than the LLM-based attribute selection approaches (\textit{Set Selection} and \textit{Scoring}) [lower mean lines of statistical baselines; \textbf{RQ2}].}
	\label{fig:results_overview}
\end{figure}

\subsection{RQ1: Impact of Human Response Variation on Persona Prompting Performance} % draft done
\label{sec:results_rq1}

Figure~\ref{fig:results_overview} shows that the persona prompting approach performs better for predicting responses to target questions with high human response variation than to questions with low human response variation across all benchmarked attribute selection approaches. For the LLM-based prediction without persona conditioning (\textit{No Persona}), this effect is reversed \textemdash the tested LLMs are better able to represent the distribution of human responses for questions with low human response variation. These two observations suggest that the inclusion of relevant persona attributes in prompts indeed allows the LLM to meaningfully model the variation in human responses. 

In Appendix~\ref{app:results_models}, we discuss the results disaggregated based on the two models families and based on the four different surveys. We show that the effect of human response variation on the predictive performance of predicting responses with and without personas generally also holds on these two levels of disaggregation, suggesting generalizability of our finding.

\subsection{RQ2: Performance Ranking of Attribute Selection Approaches} % draft done
\label{sec:results_rq2}

Figure~\ref{fig:results_overview} shows that selecting persona attributes using statistical methods (\textit{Correlation} and \textit{Feature Importance}) leads to better alignment between the LLM-predicted and the actual survey responses than using the LLM-based selection approaches (\textit{Set Selection} and \textit{Scoring}). This is interesting since one could assume that LLMs can tap into the large amount of (scientific) background knowledge that they saw during pre-training to select the (theoretically) relevant attributes, or that they leverage potential differences in how knowledge is represented to select LLM-specific attributes. Our results show that this is clearly not the case since LLM-based attribute selections do not only perform worse but are also inconsistent in their selection of attributes (see Section~\ref{sec:results_stability}).

When comparing the two LLM-based approaches against each other, we see that the set selection approach performs better than the scoring approach. In addition, using persona prompting for response prediction consistently leads to better alignment with the human responses than the LLM-baseline that does not condition on personas. This holds true across all approaches we test for the selection of persona attributes. Finally, the performance differences between the different attribute selection approaches are more pronounced for high than for low human response variation questions. We find that the performance order of attribute selection methods is consistent across model families and surveys (see Appendix~\ref{app:results_surveys}).

\subsection{Stability of LLM-based Attribute Selection Approaches}
\label{sec:results_stability}

For the LLM-based attribute selection approaches (set selection and scoring), we separately analyze the consistency with which the six LLMs choose the five most important attributes for predicting a given target question across independent runs.

\subsubsection{Stability of the Set Selection Approach}
\label{sec:stability_setselection}

\begin{figure}
    \centering
    \includegraphics[alt={Graphs comparing the stability of LLM-based set selection approach for identifying relevant persona attributes across different LLMs.}, width=\textwidth]{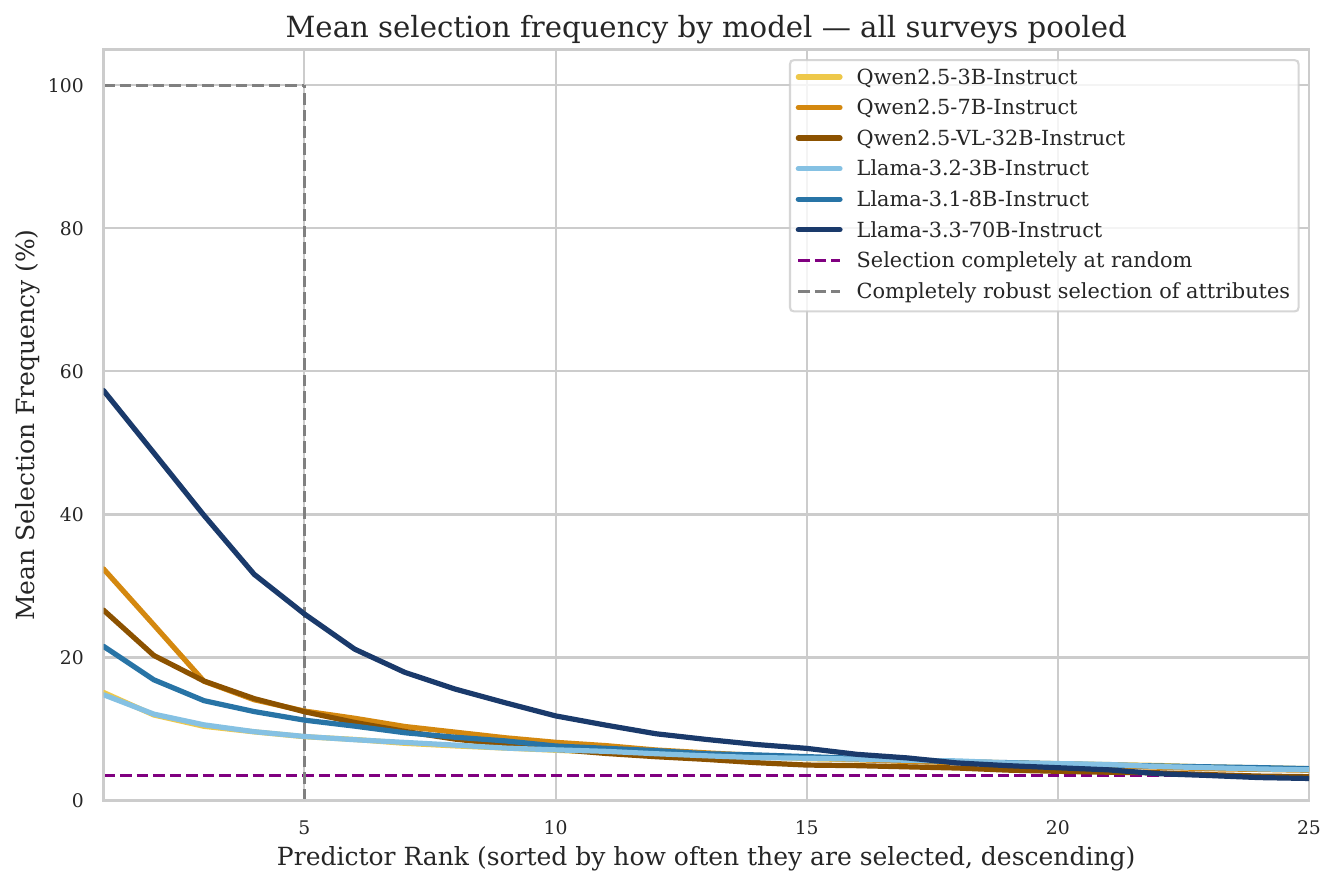}
    \caption{\textbf{The stability of the LLM-based set selection approach increases with model size.} Each curve represents the distribution of the relative frequencies (y-axis) with which the different variables in a survey (x-axis) were chosen among the top-five most important attributes for predicting a given target question in one of 100 (30 for Qwen2.5-32B and Llama3.3-70B) independent runs. The dotted blue curve represents a perfectly stable selection mechanism, selecting the same five attributes in every single run, and the dotted red line represents a selection mechanism that is random guessing, selecting a different set of variables every time. We cut the x-axis at the 25 most frequently selected variables, given that afterwards the selection frequencies approach zero. The stability of the set selection approach increases with model size, with Llama3.3-70B selecting the same attribute as most important in 58\% of the runs across target questions, with the same attribute selected as most important for some target questions in more than 80\% of the runs.}
    \label{fig:selection_robustness_mean}
\end{figure}

Figure~\ref{fig:selection_robustness_mean} shows that the stability of the selection approach increases with model size. Both for the Llama and the Qwen models, more complex LLMs are able to more consistently select the same attributes as most important. Especially the large Llama model consistently picks the same set of attributes for some target questions, with the same attribute being selected as most important in almost 60\% of runs across target questions. See Appendix~\ref{app:results_stability} for a more extensive discussion of the set selection stability results.

\subsubsection{Stability of the Scoring Approach}
\label{sec:stability_scoring}

\begin{figure}
    \centering
    \includegraphics[alt={Graphs comparing the stability of LLM-based scoring approach for identifying relevant persona attributes across different LLMs.}, width=\textwidth]{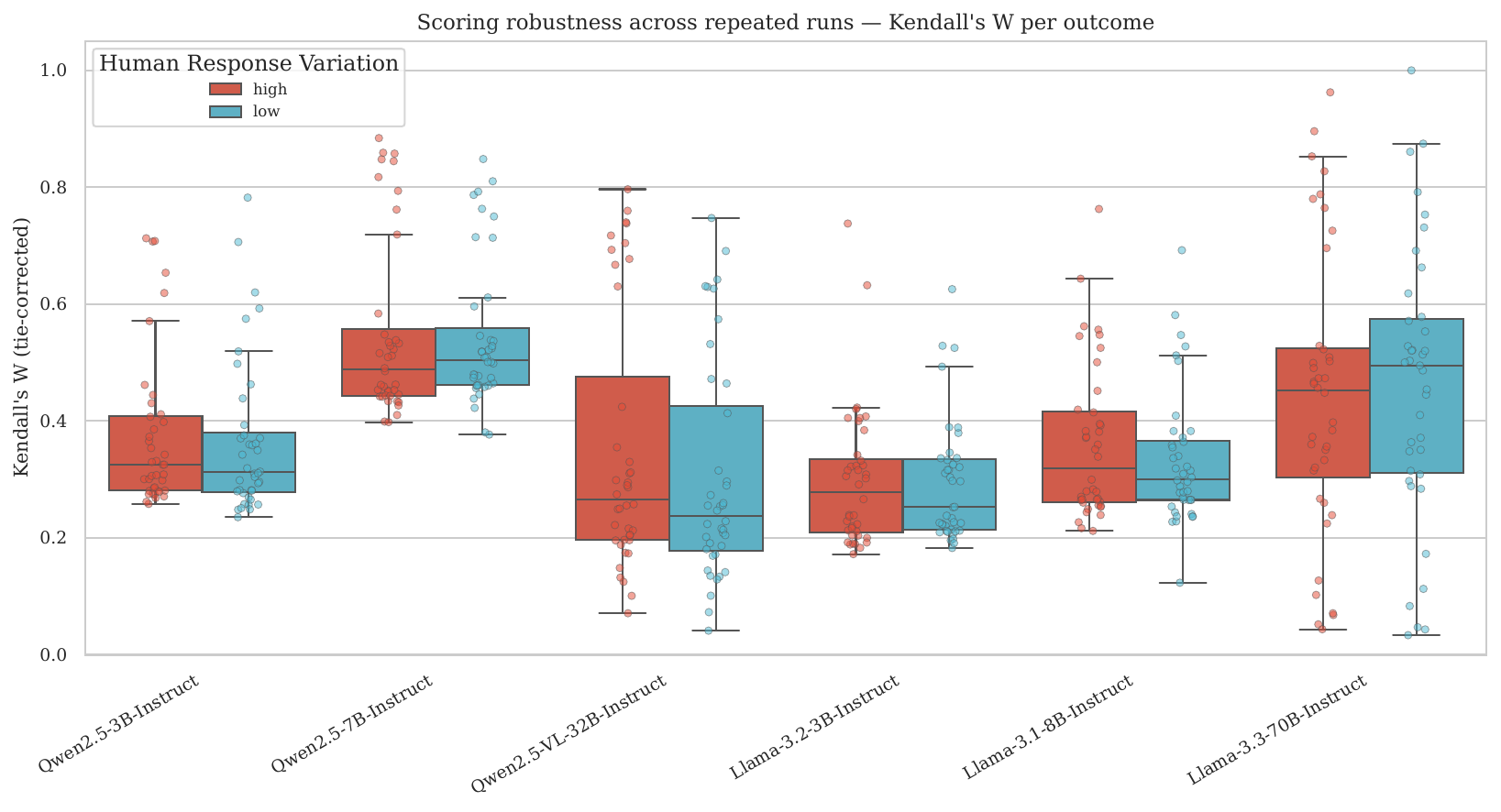}
    \caption{\textbf{LLMs struggle to reproduce the same variable importance ranking across different runs in the LLM-based scoring approach.} Each boxplot represents the distribution of the Kendall's $W$ values, a concordance correlation measure calculated from the variable importance rankings that we construct from repeated independent runs for the 20 target questions. We disaggregate results based on human response variation and the used LLM. The correlation between the importance rankings produced in independent runs is generally low. For Llama models, however, we see that the correlation increases with model size, implying higher consistency with which attributes are scored as important.}
    \label{fig:scoring_robustness_mean}
\end{figure}

Figure~\ref{fig:scoring_robustness_mean} shows that LLMs are generally unable to reproduce the importance scores assigned to predictors across multiple independent runs, with observed mean Kendall's $W$ values between 0.25 and 0.5 indicating a weak to moderate correlation between the resulting rankings. For Llama models, the stability of the importance rankings produced in independent runs generally increases with model size. This is not the case for Qwen models, for which the largest model has the lowest mean correlation coefficient across different target questions. We see that especially for larger models such as Llama-3.3-70B-Instruct, the variance in rank correlation coefficients across runs is high, indicating that the order of scores can almost perfectly be reproduced for some target variables, but fluctuates highly for others. The opposite is the case for smaller models\textemdash their variance of rank correlation coefficients is lower, indicating that regardless of the target question the importance scores generated across different runs fluctuate highly.

\section{Discussion and Implications}
\label{sec:discussion}

For practitioners interested in simulating survey responses via persona prompting, our results suggest a two-step framework for practical use. In a first step, our measures of human response variation help identify types of survey questions for which the simulation can be expected to work best. As we have shown above, this will generally be the case for questions with high high human response variation, i.e., questions (and potentially tasks and topics) that are highly contested. In a second step, our results suggest to use approaches informed by existing survey data for the same or a similar question to identify the attributes that are most important for persona creation and lead to the best alignment between predicted and actual responses. This is similar to the idea by \citet{hu2024quantifying}, who suggest to use simplistic regression analysis based on small-scale human data to estimate the likely performance of simulation efforts, enabling informed decision making before the commitment of additional resources. 

Such a simplistic framework could potentially pave the way towards an AI-based adaptive survey design framework, in which specific questions are shown only to selected groups of participants, increasing the effectiveness of the allocation of survey resources. If we know that some types of questions are particularly suitable for simulation through persona prompting and that some more majority-aligned subgroups of the population can be modeled particularly well with LLMs, then we can aspire to improve on the survey response prediction task even with the imperfect performance of current social simulation methods.

Interestingly, we observe that the statistical baseline approach which relies on pure correlations between the target variable and the input variables leads to the best selection of persona attributes and outperforms more complex approaches that rely on LLMs or semantic similarity measures. This implies that the response prediction via persona prompting LLMs requires input attributes that are correlated with the outcome variable \textemdash something that is obvious for existing methods, but much less so for LLM-based approaches. For traditional Machine Learning applications to perform well, feature selection approaches are typically designed to select features that are highly correlated with the target variable, but not highly correlated among each other. For persona prompting LLMs we do not find any evidence that correlation among selected features matters since the best performing approach simply focuses on the correlation between individual input variables and the target variable. 

One of the assumptions of using persona prompting is that LLMs are able to activate internal representations of knowledge based on the provided persona description, but so far there has been no effort to study whether the response generation is then driven by correlations and associations that can also be observed in traditionally collected human survey data, or whether it relies on LLM-specific correlations and associations between persona attributes and target variables. While our work is far from providing definite answers to this question, it serves as a first indication that the internal representations of survey responses in LLMs might be similarly correlated as in human survey responses. If that was not the case, we would expect the LLM-based selection approaches, having access to the model's internal representations, to be at an advantage and thus to perform better in predicting survey responses than survey-data-informed alternatives\textemdash which we do not observe. 

A similar and promising direction of future work would be on the inner workings of the LLM in generating its responses conditioned on a given persona (e.g., using inner probing \citep{maiya-etal-2025-improving}). While we currently control the inputs via the selection of attributes to feed into persona descriptions, we can only observe the generated outputs. Studying the parameters and connections that are actually activated during response generation and comparing them to some representations of the persona attributes provided as inputs would make for very relevant work towards forming a better understanding of the persona prompting approach. This strain of research connects to the literature on explainable AI and the concept of faithfulness in particular, which is concerned with establishing whether the explanations provided by the model, in our case the selection of persona attributes deemed important for response prediction, match the inner workings of the model \citep{madsen2024self, matton2025walk, chuang2026faithlm}. In our case, this would mean checking whether the provided persona attributes in the prompt lead to activations in the model that actually influence the predicted response.

\section*{Limitations}
\label{sec:limitations}

As is often the case in this type of work trying to fathom the frontiers of novel methods and applications, there are a number of limitations naturally arising by virtue of being incapable to consider every potentially relevant factor at once.

In our case, the list of such omitted influence factors is long, including but not limited to: the design choices of the persona attribute selection methods (always choosing five attributes instead of varying the number; trying different prompts or even entirely different LLM-based attribute selection strategies); the choice of LLMs (using SOTA closed models; using models with known training data; using base models; using different models for attribute selection and for response generation); the persona prompting response generation setup (using different persona formats; using different generation parameters; using different response parsing mechanisms). While it would be interesting to explicitly account for and systematically vary all these factors, this would further blow up an already highly compute-intensive experimental setup.

Another type of omission results less from a (computational) resource constraint but from a lack of limited capacities and attention as a researcher. For this work in particular, a number of alternative, equally relevant baselines to those we already used are perceivable, in our case especially for the attribute selection approaches. Selecting attributes fully at random or not selecting at all, instead using all attributes for persona constructions, would help to further shed light on the importance of the attribute selection step. Selecting attributes based on established social science theories would help to further improve our understanding of why persona prompting works, whether it works the way it is supposed to, and when it should be expected (not) to work. However, such a theory-informed attribute selection approach is inherently difficult since there is often no single theory for each construct and even if there was one, there is often no straightforward way of turning a theory into a agreed upon ranked lists of persona attributes. Finally, we currently lack a baseline that uses traditional survey imputation methods for the response prediction task to better contextualize the performance and to provide convincing answers to the question whether these efforts are worth our while. However, we already know from previous work that in most settings persona-prompted LLMs are competitive with a random forest classifier used as the imputation baseline alternative \citep{rupprecht2025german}.

In addition, there are some parts of this work for which additional qualitative analyses could lead to additional interesting insight. For instance, we are currently not conducting any qualitative check of and comparison across the sets of selected attributes \textemdash neither across attribute selection approach to uncover biases or preferences, nor across LLMs to identify model-specific selection patterns, nor across surveys and countries to evaluate the cultural adaptability and sensitivity to nuance of the selection methods.

Apart from these practical limitations specific to our work, there exists a growing number of position papers on the promises and limitations of LLM-based social simulations, in-silico samples, and persona prompting. \citet{anthis2025position} describe five key challenges that LLM social simulations must address in order to achieve their promise: lack of human diversity, systematic biases, sycophancy, alienness and non-human like mechanisms leading to superficially accurate results, and generalizability across contexts.

\section*{Ethical Considerations}

\citet{kirk2024benefits} provide an extensive discussion of the benefits and risks of aligning LLMs to individuals. While our main application concerns the aggregated population level, the process of generating individual-level responses via persona prompting an be viewed as analogous to aligning LLMs to individuals. One individual-level risks discussed by \citet{kirk2024benefits} that is relevant for our context is the danger of essentialism and profiling, i.e., categorizing individuals based on an insufficient set of demographic information. The risk of flattening representation and caricaturing individuals has also been raised by \citet{cheng2023marked},  \citet{cheng2023compost}, and \citet{wang2025large}. We would argue that our work is contributing to reduce risks of essentialism, since we are actively exploring ways of selecting attributes that go beyond the limited set of socio-demographic attributes usually tried first in persona prompting applications.

Another ethical consideration for the work with human participant data for developing and evaluating social simulations concerns participants' privacy, and especially the danger of collecting and potentially disclosing increased quantities of sensitive information. While we acknowledge the importance of this consideration, we argue that our work is not adding to this risk: we are only reusing already collected survey data, and we make sure to only use locally-hosted models for our experiments, avoiding the possibility of any data leakage into hosted-models to occur \citep{cheng2026privacy}.

On the societal level, relevant ethical considerations discussed by \citet{kirk2024benefits} include the displacement of labor, environmental harms through increased LLM usage, and the potential for malicious use. These considerations are and remain relevant for all LLM-based applications.

Finally, there is the question for the overarching goal of exploring the potential of social simulations and the future of survey research with human participants. We explicitly caution against the outright replacement of human survey participants through silicon samples, but at the same time advocate for keeping on exploring the potential of LLM-based applications in the survey context. Only this type of methodological evaluation and advancement allows us to make informed decisions about when and when not to use novel methods and paradigms.

\printbibliography

\newpage

% \input{appendix_a}
% \captionsetup{hypcap=false}

\section{Appendix: Prompts}
\label{app:prompts}

\subsection{Paraphrasing of Survey Questions}
\label{app:prompts_data}
% \textbf{Codebook Parsing}

In this subsection, we document the prompts used for paraphrasing the survey questions. This step was done using GPT-5.1 accessed via the OpenAI API with default parameter choices. Generated paraphrases were assessed manually. The quality and suitability of the paraphrases were found to be sufficiently high after manual validation, which is not surprising given how trivial the task was.  

\begin{promptbox}
\ttfamily\small

\textbf{User Prompt} \newline
Your task is to help translate information on survey questions into concise and self-contained paraphrases. While survey questions are optimized to elicit nuanced responses from human participants, our objective is to create key-value pairs of survey questions (key) and corresponding survey responses (value) which can be used to characterize the attributes and attitudes of the human participants. \newline

The resulting key-value pairs will be used as persona descriptions to steer the generation of LLMs in a persona prompting setup. You will be given a list of survey questions. For each question, there is a short and a long version of the name of the question, a detailed description (often containing instructions for the interviewer), as well as the response options. \newline

You are tasked to create a paraphrase of the survey question that complies with the following requirements: 

\begin{itemize}
    \item concise: the paraphrase should omit any unnecessary details such as interviewer instructions 
    \item self-contained: the paraphrase should not rely on the context of the survey or the presence of the response options to make sense
    \item consistent: the paraphrases should be consistent across similar survey questions. In surveys, there are often batteries of questions that only differ in two or three words - the paraphrases of these questions should also only differ in these two or three words
    \item matching response options: the paraphrases should be written such that the available response options match the paraphrase
\end{itemize}

In the list of survey questions, each survey question is represented by a JSON object with keys "name\_short", "name\_long", "description", and "responses". As output, generate a list of paraphrases for these survey questions. Each entry of that list should correspond to one paraphrased survey question, should be formatted as a JSON object, and should have the keys "name\_short" and "paraphrase". Return only the list of JSON objects in your response. \newline

Here is the list of survey questions: <list\_survey\_items> 

\end{promptbox}
\captionof{figure}{\textbf{User prompts used for the paraphrasing of survey questions.} Placeholders are marked as <placeholder> and are replaced with the respective information.}
\label{fig:prompt_paraphrasing}

\begin{promptbox}
\ttfamily\small

\textbf{User Prompt} \newline

You are doing this task in batches. In addition, you are given the paraphrases you generated for the last five survey questions from the previous batch. If they belong to the same battery of questions you are processing now, please make sure that the paraphrases are consistent with the previous batch. 

These are the last five paraphrases you generated: <last\_five\_paraphrases> 

\end{promptbox}
\captionof{figure}{\textbf{User prompts used for the batch paraphrasing of survey questions.} This prompt is appended to the base paraphrasing prompt for all but the first batch.}
\label{fig:prompt_paraphrasing_batch}

\subsection{Attribute Selection}
\label{app:prompts_selection}

In this subsection, we document the prompts used for the two LLM-based approaches used for persona attribute selection: set selection and scoring.

% \subsubsection{Set Selection}
% \label{app:prompts_selection_set}
% In this subsection, we document the prompts used for the LLM-based set selection approach for persona attribute selection.

\begin{promptbox}
\ttfamily\small

\textbf{System Prompt} \newline
Your job is to identify the most important survey items for predicting the responses to a specific target survey question using an LLM in a zero-shot persona-prompting setup. The identified items will be used to construct short persona descriptions, which will then be used to prompt an LLM to predict the survey response of the corresponding human participant. The survey for which you perform this job was collected in <country> in the year <year>.\newline

You will be shown the target question together with a list of all available survey items. Your task is to select the five most important survey items for predicting the individual-level responses to the target survey question in the persona-prompting setup. As the output of your reasoning, respond ONLY with a valid JSON object, formatted like in this example: \newline

\begin{Verbatim}[fontsize=\small]
{
    "reasoning":        "brief 1-2 sentence explanation",
    "selected_items":   ["id_01","id_23","id_49",“id_517”,“id_521”]
}
\end{Verbatim}

\textbf{User Prompt} \newline
Target survey question: <target\_question>
Available survey items: <list\_survey\_items> \newline

Select the five most important survey items for predicting the individual-level responses to the target survey question. Respond with the correct JSON object format. 

\end{promptbox}
\captionof{figure}{\textbf{System and user prompts used for the \emph{Set Selection} persona attribute selection approach.}}
\label{fig:prompt_set_selection}

% \subsubsection{Scoring}
% \label{app:prompts_selection_scoring}
% In this subsection, we document the prompts used for the LLM-based scoring approach for persona attribute selection.

\begin{promptbox}
\ttfamily\small

\textbf{System Prompt} \newline
Your job is to identify the most important survey items for predicting the responses to a specific target survey question using an LLM in a zero-shot persona-prompting setup. The identified items will be used to construct short persona descriptions, which will then be used to prompt an LLM to predict the survey response of the corresponding human participant. The survey for which you perform this job was collected in <country> in the year <year>. \newline

You will be shown the target survey question together with each available survey item individually. Your task is to score each of them according to their importance for predicting the individual-level responses to the target survey question in the persona-prompting setup, with possible integer scores ranging from 0 (not important at all) to 100 (most important). As the output of your reasoning, respond ONLY with a valid JSON object, formatted like in this example: \newline

\begin{Verbatim}[fontsize=\small]
{
    "reasoning":        "brief 1-2 sentence explanation",
    "importance_score": [0-100]
}
\end{Verbatim}

\textbf{User Prompt} \newline
Target survey question: <target\_question> 
Survey item: <survey\_item> \newline

Assign an integer importance\_score between 0 (not important at all) and 100 (most important) to the survey item. 

\end{promptbox}
\captionof{figure}{\textbf{System and user prompts used for the \emph{Scoring} persona attribute selection approach.}}
\label{fig:prompt_scoring}

\subsection{Response Prediction}
\label{app:prompts_prediction}

We generate synthetic survey response distributions for each target question by prompting the LLMs with the persona descriptions constructed from the most important attributes. Each survey participant's response is generated via a separate prompt. LLMs are restricted to generate a reasoning stream and finalize their answer with a dictionary format response including a short explanation of response choice and the final response.

\subsubsection{Persona Prompting Approaches}
\label{app:prompts_prediction_persona}

\begin{promptbox}
\ttfamily\small

\textbf{System Prompt} \newline

You are simulating a survey respondent, participating in a survey in <country> in the year <year>.  You will be given a profile of a person described by their answers to several survey questions. Your task is to predict how this person would answer a specific survey question, choosing from the provided answer options. Respond ONLY with a valid JSON object: \newline

\begin{Verbatim}[fontsize=\small]
{
    "reasoning":            "brief 1-2 sentence explanation grounded 
                            in the profile",
    "predicted_response":   "the exact integer answer option label 
                            chosen from the answer options"
}
\end{Verbatim}

\textbf{User Prompt} \newline
You are a survey respondent, described through the following attributes:<persona\_description> \newline
What is your answer to the following question:  
<outcome\_question> \newline
These are the available answer options: 
<answer\_options> \newline
Predict the most likely answer for this respondent. Respond ONLY with a valid JSON object. 

\end{promptbox}
\captionof{figure}{\textbf{System and user prompts used for the \emph{persona prompting} response prediction approaches.}}
\label{fig:prompt_response_prediction}

\subsubsection{No Persona Baseline}
\label{app:prompts_prediction_nopersona}

\begin{promptbox}
\ttfamily\small

\textbf{System Prompt} \newline

You are simulating a survey respondent, participating in a survey in <country> in the year <year>. Your task is to reason and predict a survey respondent's answer by choosing from the provided answer options. Respond ONLY with a valid JSON object: \newline

\begin{Verbatim}[fontsize=\small]
{
    "reasoning":            "brief 1-2 sentence explanation of 
                            your choice",
    "predicted_response":   "the exact integer answer option label 
                            chosen from the answer options"
}
\end{Verbatim}

\textbf{User Prompt} \newline
You are a survey respondent. \newline
What is your answer to the following question:  
<outcome\_question> \newline
These are the available answer options: 
<answer\_options> \newline
Reason and predict the most likely answer for the respondent. Respond ONLY with a valid JSON object. 

\end{promptbox}
\captionof{figure}{\textbf{System and user prompts used for the \emph{no persona} baseline for response prediction.}}
\label{fig:prompt_response_prediction_nopersona}

\section{Appendix: Experimental Setup}
\label{app:experimental_setup}

\subsection{Survey Data}
\label{app:experimental_setup_data}

In the following, we document the preprocessing steps for preparing the survey data for our persona prompting experiments, as well as the codebooks used for the annotation of the scale and type of the survey variables.

For all surveys, we use the latest publicly available data release. To prepare the survey data for use in our experiments, a number of pre-processing steps are necessary, mostly transforming and standardizing the format of the survey questions and responses. Most surveys still rely on pdf codebooks to make their data accessible and legible to researchers. While this might work for more traditional survey applications, this makes the handling and use of the data at the scale required for persona applications burdensome. Thus, the first step is to bring the codebook, and particular the texts of the survey questions and the available response options, into machine-readable formats. Where necessary, we create custom codebook-parsers to extract the relevant information in a structured way.

Once the codebooks have been parsed, we annotate the survey variables for their scale (nominal, ordinal) as well as for the type of information they represent (socio-demographic, behavioral/circumstantial, attitudinal, relational, contextual, non-participant). See Tables \ref{tab:codebook_scale} and \ref{tab:codebook_type} for the respective annotation codebooks. We require the scale of a variable for deciding which measure of disagreement is appropriate, and the type of information to identify the subsets of survey variables considered relevant for persona construction and evaluation. In addition, we identify and remove variables that can be considered direct duplicates of each other. 

\begin{table}[!ht]
    \renewcommand{\arraystretch}{1.5} % increase spacing between table-rows
    \centering
    \begin{tabular}{p{2.5cm}p{12.5cm}}
        \multicolumn{2}{l}{\textbf{Variable Scale}} \\
        Nominal &  All survey variables with nominal response options, e.g., binary "yes"/"no" or "marked"/"unmarked" choices, as well as categories without clear order or clearly defined extreme values, e.g., party affiliation.\\
        Ordinal &  All survey variables with ordinal response options, i.e., with clear order and clearly defined extreme values, e.g., Likert scales, age (measured in years), or year of birth.\\
        Continuous & All survey variables with continuous response scales, i.e., an open-ended, numeric response field. This mainly applies to variables for survey weights or different variants of income.\\
        \\
    \end{tabular}
    \caption{\textbf{Annotation codebook for the scale of survey variables.} We annotate the scale of the survey variables in order to calculate the appropriate measure of human response variation.}
    \label{tab:codebook_scale}
\end{table}

\begin{table}
    \renewcommand{\arraystretch}{1.5} % increase spacing between table-rows
    \centering
    \begin{tabular}{p{2.5cm}p{12.5cm}}
        \multicolumn{2}{l}{\textbf{Variable Type}} \\
        Socio-Demographic &  We annotate all variables that concern one of the following aspects as socio-demographic: age, gender, sex, income, wealth, race/ethnicity, origin, migration status, and education. In addition, we include "core" job-related variables, both for the current and the previous job: employment status, job title, and time in job.\\
        Behavioral/ Circumstantial &  In this category, we include all variables that concern past (e.g., "Have you smoked in the last 30 days?") or future (e.g., "Do you plan to participate in upcoming elections?") behavior, as well as customary behavior (e.g., "Do you regularly attend church?"). In addition, we include circumstances (e.g., "Rate your general health", "Have you experienced bullying when growing up?") and general sentiments and feelings (e.g., "Do you feel depressed?", "How satisfied are you with your life?"). Additional details on the participant's employment are also collected in this category, e.g., weekly working hours, possible risk factors at work, or personnel responsibility.\\
        Attitudinal & In this category, we include all variables that measure the participant's opinions, attitudes, sentiments, and beliefs. This includes many questions starting with "In your opinion ..." or "Do you agree/disagree ...". This category also includes questions on the participant's religious, political, and ideological beliefs.\\
        Relational & In this category, we include all variables that are not on the survey participant in isolation, but on their relationship to other members of their family or household, or characteristics and attributes of these other members. These other members typically are the participant's spouse or partner, their parents, their children, or other members of their household.\\
        Contextual & In this category, we include a small set of variables that are immediately relevant for the survey context. They include the participant's ID, as well as the country and year in which the survey was conducted.\\
        Non-Participant & Variables that are not measuring attributes of the participant. These include technical variables such as weights or flags, questions about the interview or attributes of the interviewer, as well as the interviewer's perception of the participant.\\
        \\
    \end{tabular}
    \caption{\textbf{Annotation codebook for the type of survey variables.} We annotate the type of the survey variable in order to only select relevant sets of survey variables for the selection of persona attributes and target variables.}
    \label{tab:codebook_type}
\end{table}

\newpage

\subsection{Target Questions}
\label{app:experimental_setup_targets}

\begin{figure}
    \centering
    \includegraphics[width=\linewidth]{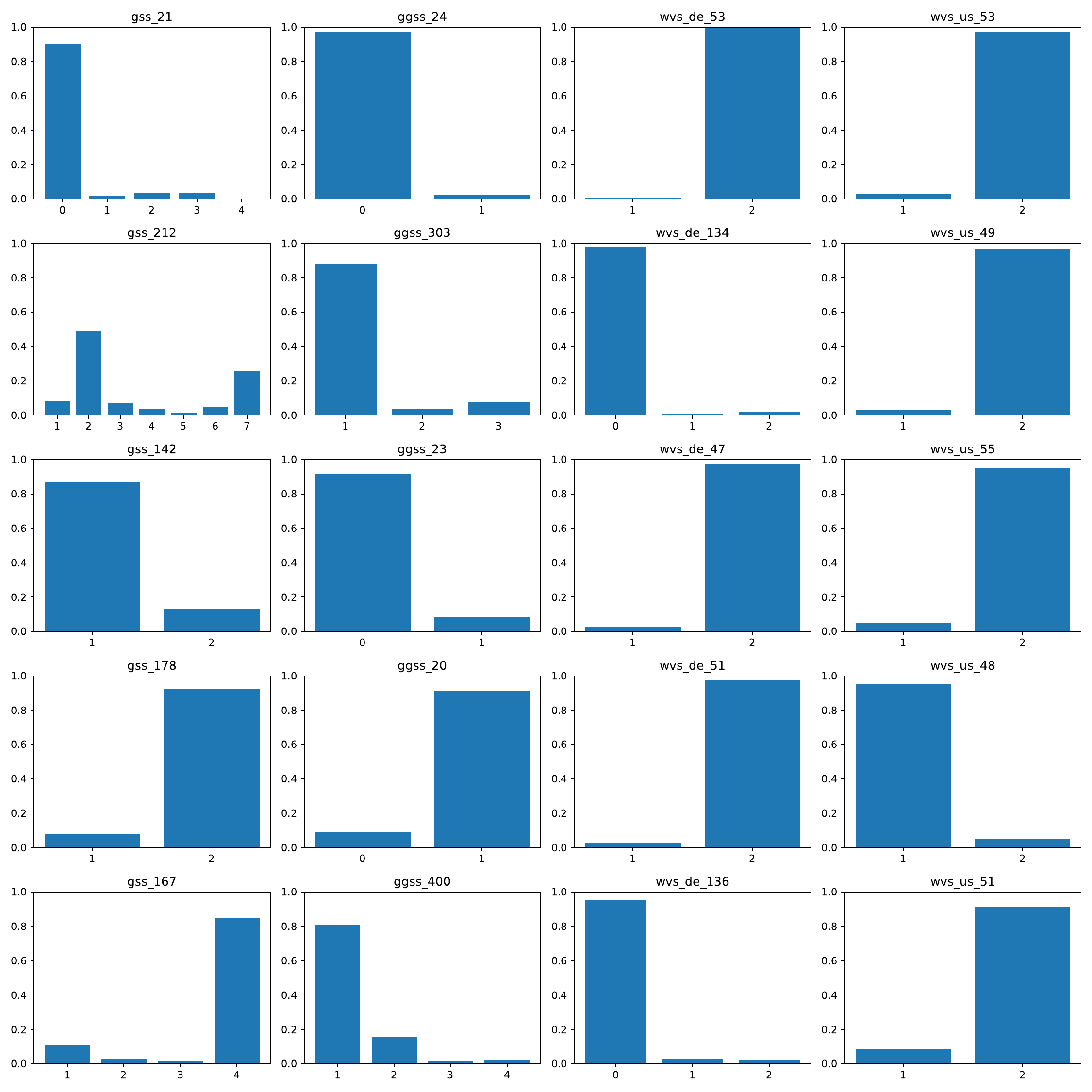}
    \caption{Human response distribution of the five \textbf{nominal} target variables with \textbf{lowest human response variation}. Each column represents target questions selected for one of the four surveys, with human response variation increasing from top to bottom.}
    \label{fig:targets_nominal_low_HRV}
\end{figure}

\begin{figure}
    \centering
    \includegraphics[width=\linewidth]{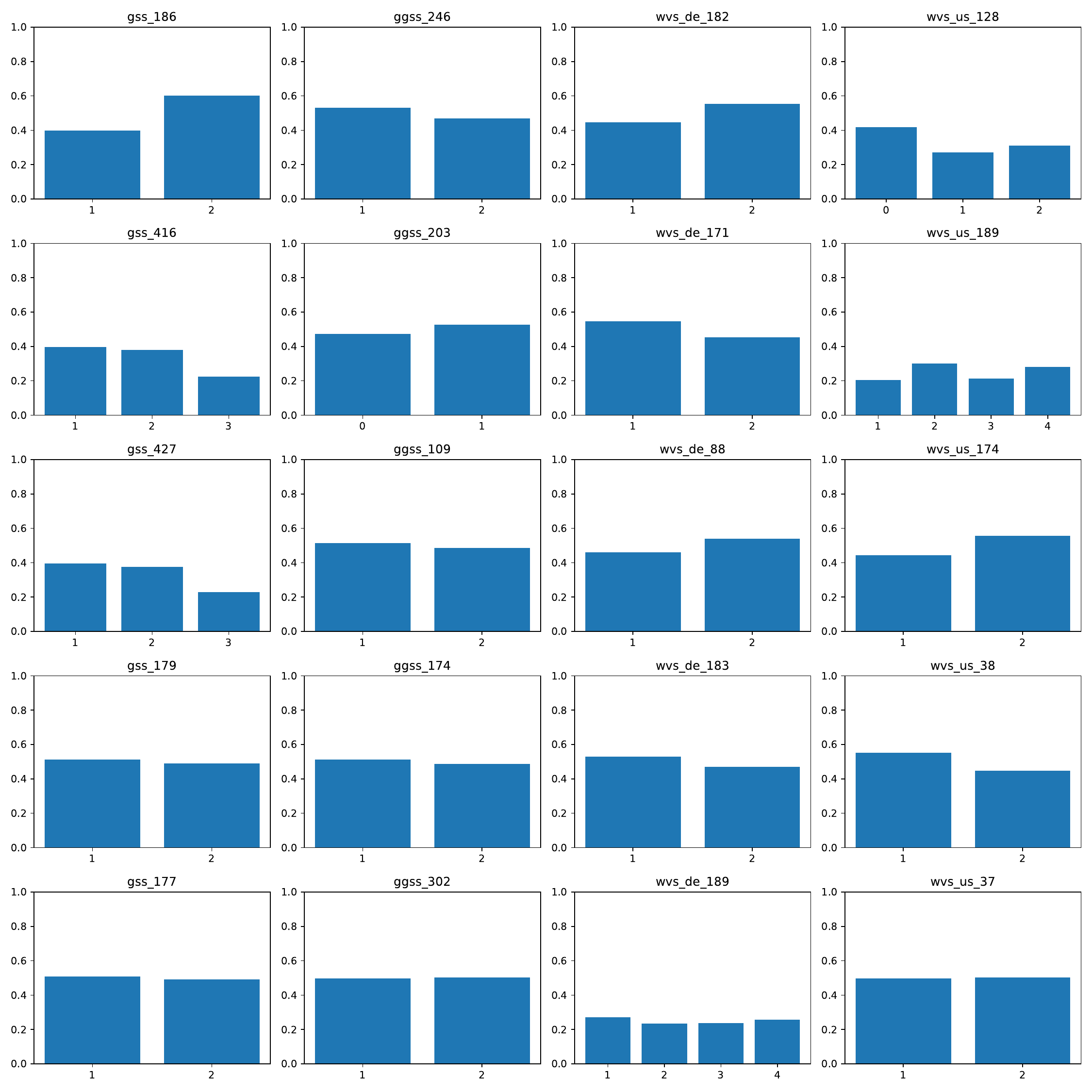}
    \caption{Human response distribution of the five \textbf{nominal} target variables with \textbf{highest human response variation}.}
    \label{fig:targets_nominal_high_HRV}
\end{figure}

\begin{figure}
    \centering
    \includegraphics[width=\linewidth]{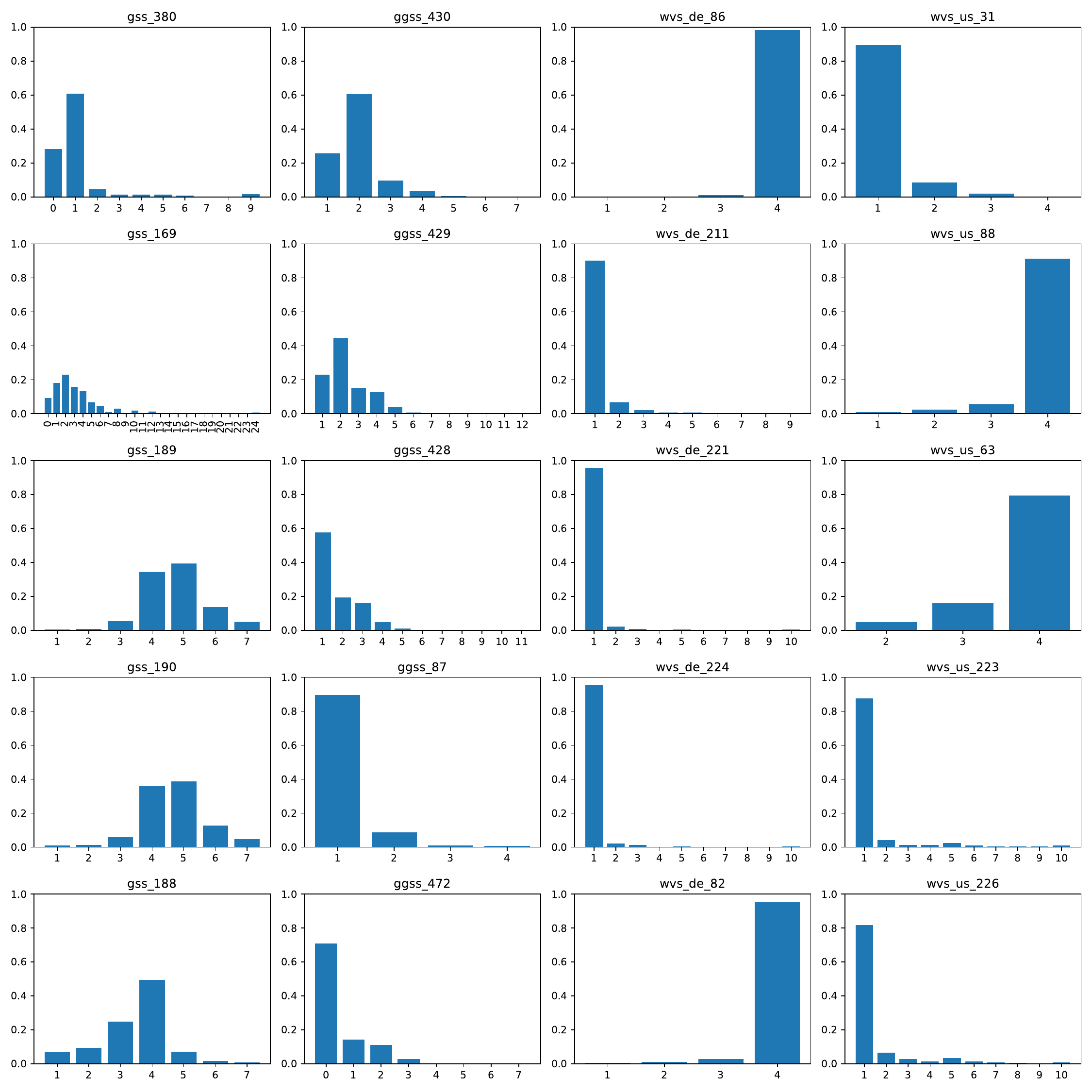}
    \caption{Human response distribution of the five \textbf{ordinal} target variables with \textbf{lowest human response variation}.}
    \label{fig:targets_ordinal_low_HRV}
\end{figure}

\begin{figure}
    \centering
    \includegraphics[width=\linewidth]{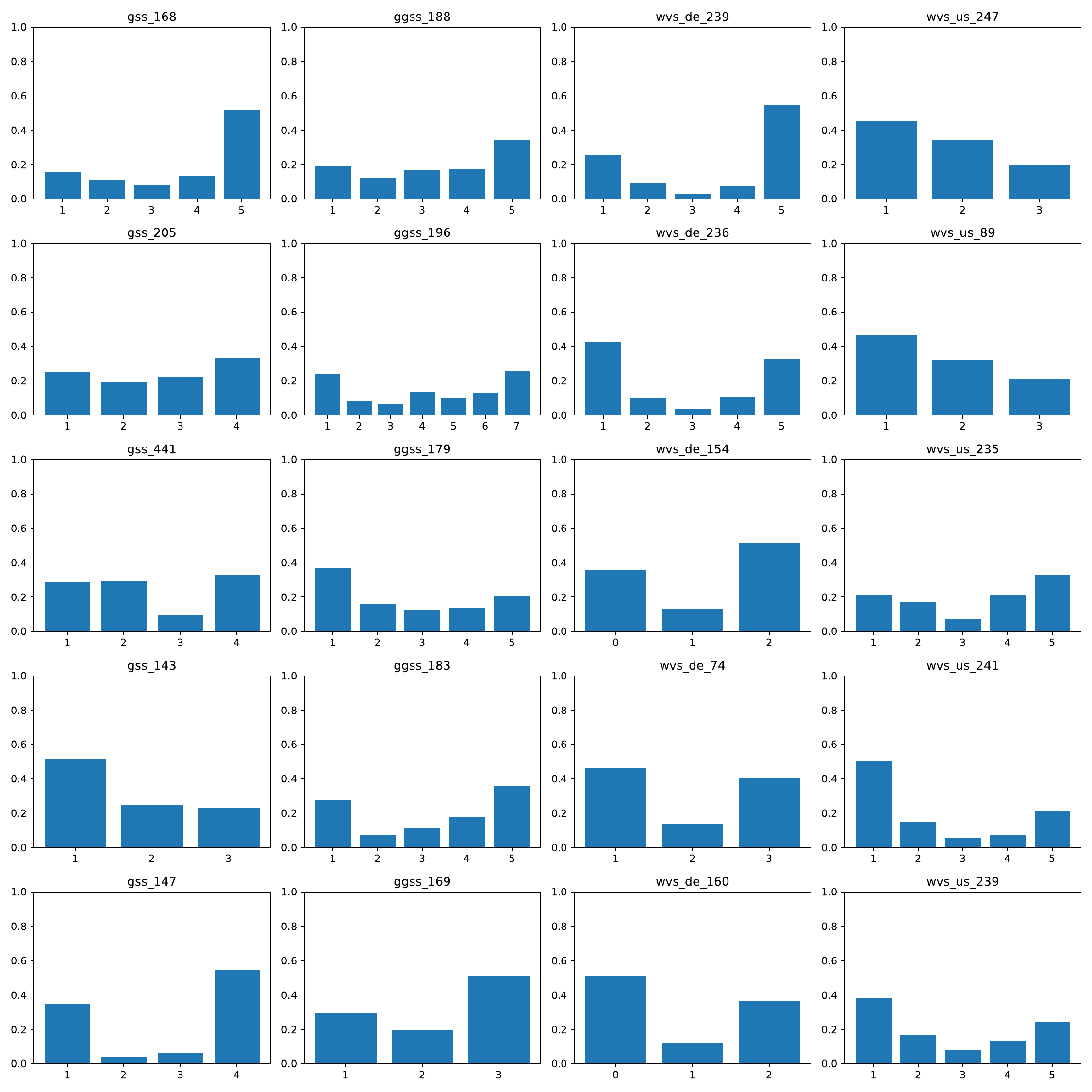}
    \caption{Human response distribution of the five \textbf{ordinal} target variables with \textbf{highest human response variation}.}
    \label{fig:targets_ordinal_high_HRV}
\end{figure}

Figures~\ref{fig:targets_nominal_low_HRV} to \ref{fig:targets_ordinal_high_HRV} show the human response distributions of the 20 target variables selected for each survey. The figures are separated into variables with nominal and ordinal scale, as well as into low and high response variation. Tables~\ref{tab:target_questions_gss} to \ref{tab:target_questions_wvs_us} provide the texts of those target questions, as well as their human response variation and the number of valid responses. The variable ID (Var-ID) provided both in the figures and in the tables serve as a link between the two sources of information.

\begin{table}
    \footnotesize
    \renewcommand{\arraystretch}{1.2} % increase spacing between table-rows
    \begin{tabular}{p{1.5cm}p{11cm}crr}
        Var-ID & Question Text & Type & HRV & n \\
        \bottomrule
        \textit{Nominal} & \textit{Low Human Response Variation} \\
        \midrule
        gss\_21 & How much time, if any, have you spent on active-duty military service? & BC & 0.2596 & 3292 \\
        gss\_212 & What is the first other language you speak besides English (or Spanish)? & BC & 0.2706 & 847 \\
        gss\_142 & Do you favor or oppose sex education being taught in public schools? & AT & 0.3492 & 2131 \\
        gss\_178 & Do you think Black–white differences in average jobs, income, and housing are because most Black people have less in-born ability to learn? & AT & 0.3892 & 2122 \\
        gss\_167 & Do you or your spouse go hunting, and if so, who hunts? & BC & 0.3988 & 2229 \\
        \midrule
        \textit{Nominal} & \textit{High Human Response Variation} \\
        \midrule
        gss\_186 & Have you had a “born again” experience, meaning a turning point when you committed yourself to Christ? & BC & 0.9695 & 3185 \\
        gss\_416 & Do you think most people would try to take advantage of you if they could, or would they try to be fair? & AT & 0.9742 & 1221 \\
        gss\_427 & Do most people try to be helpful, or are they mostly just looking out for themselves? & AT & 0.9755 & 1232 \\
        gss\_179 & Do you think Black–white differences in average jobs, income, and housing are because most Black people lack the educational opportunities needed to rise out of poverty? & AT & 0.9996 & 2099 \\
        gss\_177 & Do you think Black–white differences in average jobs, income, and housing are mainly due to discrimination against Black people? & AT & 0.9998 & 2085 \\
        \bottomrule
        \textit{Ordinal} & \textit{Low Human Response Variation} \\
        \midrule
        gss\_380 & How many sexual partners have you had in the last 12 months? & BC & 0.1106 & 2041 \\
        gss\_169 & On a typical day, about how many hours do you personally watch television? & BC & 0.1523 & 2152 \\
        gss\_189 & On a scale where 1 means almost all rich and 7 means almost all poor, how would you rate Blacks in general? & AT & 0.2178 & 1068 \\
        gss\_190 & On a scale where 1 means almost all rich and 7 means almost all poor, how would you rate Hispanic Americans in general? & AT & 0.2234 & 1061 \\
        gss\_188 & On a scale where 1 means almost all rich and 7 means almost all poor, how would you rate whites in general? & AT & 0.2380 & 1067 \\
        \midrule
        \textit{Ordinal} & \textit{High Human Response Variation} \\
        \midrule
        gss\_168 & How often do you read a newspaper? & BC & 0.6737 & 2168 \\
        gss\_205 & Do you favor or oppose giving women preference in hiring and promotion because of past discrimination, and how strongly? & AT & 0.6825 & 1058 \\
        gss\_441 & Would you describe your religious affiliation as strong, not very strong, somewhat strong, or having no religion? & AT & 0.7124 & 1799 \\
        gss\_143 & Should it be easier, more difficult, or about as it is now to obtain a divorce in this country? & AT & 0.7330 & 2080 \\
        gss\_147 & How wrong, if at all, do you think it is for two adults of the same sex to have sexual relations? & AT & 0.8882 & 2125 \\
        \bottomrule
        \vspace{1mm}
    \end{tabular}
    \caption{\textbf{Target variables selected for the \textbf{GSS}.} We show the paraphrased survey question texts of the five nominal and ordinal target variables with lowest and highest human response variation scores, respectively. Within the nominal and ordinal variables, variables are ordered in increasing order of their human response variation scores.}
    \label{tab:target_questions_gss}
\end{table}

\begin{table}
    \footnotesize
    \renewcommand{\arraystretch}{1.2} % increase spacing between table-rows
    \begin{tabular}{p{1.5cm}p{11cm}crr}
        Var-ID & Question Text & Type & HRV & n \\
        \bottomrule
        \textit{Nominal} & \textit{Low Human Response Variation} \\
        \midrule
        ggss\_24 & In the last three months, have you used the Internet for private purposes with any other devices? & BC & 0.1720 & 4799 \\
        ggss\_303 & Are you afraid that in the near future you might become unemployed or have to change your current job? & AT & 0.3957 & 2634 \\
        ggss\_23 & In the last three months, have you used the Internet for private purposes with an e-book reader? & BC & 0.4160 & 4799 \\
        ggss\_20 & In the last three months, have you used the Internet for private purposes with a smartphone? & BC & 0.4295 & 4799 \\
        ggss\_400 & When you were 15 years old, did you live in the same household with both of your parents? & BC & 0.4401 & 5188 \\
        \midrule
        \textit{Nominal} & \textit{High Human Response Variation} \\
        \bottomrule
        ggss\_246 & In your view, does everyone in Germany today have the opportunity to get an education that fully matches their talents and abilities? & AT & 0.9971 & 3556 \\
        ggss\_203 & If you wanted to influence politics on an issue important to you, would you consider boycotting or deliberately buying certain products for political, ethical, or environmental reasons? & BC & 0.9981 & 3417 \\
        ggss\_109 & Do you personally have contact with foreigners living in Germany in your neighborhood? & BC & 0.9994 & 3591 \\
        ggss\_174 & Have you ever previously been a member of a church or religious community? & BC & 0.9996 & 2448 \\
        ggss\_302 & Does your main job include supervising the work of other employees or telling them what they have to do? & BC & 1.0000 & 2965 \\
        \bottomrule
        \textit{Ordinal} & \textit{Low Human Response Variation} \\
        \midrule
        ggss\_430 & How many adults who belong to the survey’s target population currently live in your household, including yourself? & BC & 0.1294 & 4988 \\
        ggss\_429 & Including yourself, how many people in total currently live in your household? & BC & 0.1306 & 5144 \\
        ggss\_428 & How many additional people, apart from yourself, currently live in your household? & BC & 0.1333 & 3963 \\
        ggss\_87 & How bad or not bad do you personally consider it if a man forces his wife to have sexual intercourse? & AT & 0.1405 & 3585 \\
        ggss\_472 & How many of your own biological, step-, adoptive, or foster children live in your household? & BC & 0.1561 & 5074 \\
        \midrule
        \textit{Ordinal} & \textit{High Human Response Variation} \\
        \midrule
        ggss\_188 & How important was the reason “I left the church because I can no longer do anything with faith” for your decision to leave the church? & BC & 0.6505 & 1130 \\
        ggss\_196 & To what extent do you agree with the statement: “I would have no objection to a Muslim mayor in my municipality”? & AT & 0.6615 & 3409 \\
        ggss\_179 & How important was the reason “I left the church because I was angry with pastors or other church staff” for your decision to leave the church? & BC & 0.6866 & 1120 \\
        ggss\_183 & How important was the reason “I left the church because I can believe without the church” for your decision to leave the church? & BC & 0.7315 & 1129 \\
        ggss\_169 & What is your opinion about religious education in public schools in Germany? & AT & 0.7982 & 5107 \\
        \bottomrule
        \vspace{1mm}
    \end{tabular}
    \caption{\textbf{Target variables selected for the \textbf{GGSS}.}}
    \label{tab:target_questions_ggss}
\end{table}

\begin{table}
    \footnotesize
    \renewcommand{\arraystretch}{1.2} % increase spacing between table-rows
    \begin{tabular}{p{1.5cm}p{11cm}crr}
        Var-ID & Question Text & Type & HRV & n \\
        \bottomrule
        \textit{Nominal} & \textit{Low Human Response Variation} \\
        \midrule
        wvs\_de\_53 & Would you object to having unmarried couples living together as neighbors? & AT & 0.0571 & 1523 \\
        wvs\_de\_134 & Are you an active member, inactive member, or not a member of a consumer organization? & BC & 0.1020 & 1524 \\
        wvs\_de\_47 & Would you object to having people of a different race as neighbors? & AT & 0.1821 & 1523 \\
        wvs\_de\_51 & Would you object to having people of a different religion as neighbors? & AT & 0.1855 & 1523 \\
        wvs\_de\_136 & Are you an active member, inactive member, or not a member of a women’s group? & BC & 0.2004 & 1523 \\
        \midrule
        \textit{Nominal} & \textit{High Human Response Variation} \\
        \midrule
        wvs\_de\_182 & If you had to choose, which is more important to you: freedom or security? & AT & 0.9919 & 1476 \\
        wvs\_de\_171 & Have you avoided carrying much money for reasons of personal security? & BC & 0.9935 & 1514 \\
        wvs\_de\_88 & Generally speaking, do you think most people can be trusted, or do you think you must be very careful when dealing with people? & AT & 0.9953 & 1482 \\
        wvs\_de\_183 & If there were a war, would you be willing to fight for your country? & AT & 0.9975 & 1375 \\
        wvs\_de\_189 & Which of the following is second most important to you: a stable economy, a more humane and less impersonal society, a society where ideas matter more than money, or the fight against crime? & AT & 0.9987 & 1492 \\
        \bottomrule
        \textit{Ordinal} & \textit{Low Human Response Variation} \\
        \midrule
        wvs\_de\_86 & In the last 12 months, how often have you or your family gone without a safe place to live? & BC & 0.0262 & 1525 \\
        wvs\_de\_211 & To what extent is stealing property justifiable, if at all? & AT & 0.0518 & 1528 \\
        wvs\_de\_221 & To what extent is it justifiable, if at all, for a man to beat his wife? & AT & 0.0551 & 1527 \\
        wvs\_de\_224 & To what extent is terrorism for political, ideological or religious ends justifiable, if at all? & AT & 0.0590 & 1525 \\
        wvs\_de\_82 & In the last 12 months, how often have you or your family gone without enough food to eat? & BC & 0.0961 & 1527 \\
        \midrule
        \textit{Ordinal} & \textit{High Human Response Variation} \\
        wvs\_de\_239 & How often do you use social media to get information about what is happening in the country and the world? & BC & 0.8158 & 1527 \\
        wvs\_de\_236 & How often do you use your mobile phone to get information about what is happening in the country and the world? & BC & 0.8370 & 1526 \\
        wvs\_de\_154 & Do you agree or disagree that immigrants help your country by filling useful jobs in the workforce? & AT & 0.8644 & 1490 \\
        wvs\_de\_74 & If less importance were placed on work in people’s lives, would you see this as good, bad, or would you not mind? & AT & 0.8676 & 1477 \\
        wvs\_de\_160 & Do you agree or disagree that immigrants harm your country by increasing unemployment? & AT & 0.8770 & 1488 \\
        \bottomrule
        \vspace{1mm}
    \end{tabular}
    \caption{\textbf{Target variables selected for the \textbf{WVS\_DE}.}}
    \label{tab:target_questions_wvs_de}
\end{table}

\begin{table}
    \footnotesize
    \renewcommand{\arraystretch}{1.2} % increase spacing between table-rows
    \begin{tabular}{p{1.5cm}p{10.5cm}crr}
        Var-ID & Question Text & Type & HRV & n \\
        \bottomrule
        \textit{Nominal} & \textit{Low Human Response Variation} \\
        \midrule
        wvs\_us\_53 & Would you object to having people of a different religion as neighbors? & AT & 0.1818 & 2435 \\
        wvs\_us\_49 & Would you object to having people of a different race as neighbors? & AT & 0.2045 & 2435 \\
        wvs\_us\_55 & Would you object to having unmarried couples living together as neighbors? & AT & 0.2727 & 2435 \\
        wvs\_us\_48 & Would you object to having drug addicts as neighbors? & AT & 0.2781 & 2435 \\
        wvs\_us\_51 & Would you object to having immigrants or foreign workers as neighbors? & AT & 0.4294 & 2435 \\
        \midrule
        \textit{Nominal} & \textit{High Human Response Variation} \\
        \midrule
        wvs\_us\_128 & Are you an active member, inactive member, or not a member of a church or religious organization? & BC & 0.9843 & 2581 \\
        wvs\_us\_189 & Which of the following national aims is second most important to you: maintaining order, giving people more say in government, fighting rising prices, or protecting freedom of speech? & AT & 0.9902 & 2501 \\
        wvs\_us\_174 & Have you chosen not to go out at night for reasons of personal security? & BC & 0.9908 & 2576 \\
        wvs\_us\_38 & Do you consider independence an especially important quality for children to learn at home? & AT & 0.9918 & 2592 \\
        wvs\_us\_37 & Do you consider good manners an especially important quality for children to learn at home? & AT & 1.0000 & 2592 \\
        \bottomrule
        \textit{Ordinal} & \textit{Low Human Response Variation} \\
        \midrule
        wvs\_us\_31 & How important is family in your life? & AT & 0.1397 & 2593 \\
        wvs\_us\_88 & In the last 12 months, how often have you or your family gone without a safe place to live? & BC & 0.1576 & 2586 \\
        wvs\_us\_63 & How much do you agree that when jobs are scarce, men should have more right to a job than women? & AT & 0.1614 & 2591 \\
        wvs\_us\_223 & To what extent is it justifiable, if at all, for a man to beat his wife? & AT & 0.1643 & 2570 \\
        wvs\_us\_226 & To what extent is terrorism for political, ideological or religious ends justifiable, if at all? & AT & 0.1940 & 2570 \\
        \midrule
        \textit{Ordinal} & \textit{High Human Response Variation} \\
        \midrule
        wvs\_us\_247 & Have you donated to a group or campaign, might you do it, or would you never do it? & BC & 0.6601 & 2563 \\
        wvs\_us\_89 & Compared with your parents when they were about your age, is your standard of living better, worse, or about the same? & AT & 0.6795 & 2581 \\
        wvs\_us\_235 & How often do you use daily newspapers to get information about what is happening in the country and the world? & BC & 0.6897 & 2566 \\
        wvs\_us\_241 & How often do you use social media to get information about what is happening in the country and the world? & BC & 0.7271 & 2562 \\
        wvs\_us\_239 & How often do you use email to get information about what is happening in the country and the world? & BC & 0.7345 & 2561 \\
        \bottomrule
        \vspace{1mm}
    \end{tabular}
    \caption{\textbf{Target variables selected for the \textbf{WVS\_US}.}}
    \label{tab:target_questions_wvs_us}
\end{table}

\newpage

\subsection{Model Selection}
\label{app:experimental_setup_models}

Table~\ref{tab:model_releases} shows the dates of knowledge-cutoff and release for the LLMs used in our experiments. While we can practically rule out the possibility that GSS and GGSS has leaked into any of the used LLMs' training datasets, this would theoretically and based on the timing alone be possible for the two WVS datasets, both of which have been released 20/07/2020, well before the development of our LLMs. However, the complex storage of the data, distributed across various .csv and .pdf files, the lack of indicators of "valuable text data" contained in these files, as well as the comparability of the predictive performance for the two WVS surveys and the impossibly leaked other two surveys suggest that no actual data leakage into the LLMs has occurred.

\begin{table}[!ht]
    \centering
    \begin{tabular}{lccr}
        \hline
        \textbf{Model} & \textbf{K-Cutoff} & \textbf{R-Date} & \textbf{Huggingface Model ID} \\
        \hline
        \textsc{Qwen2.5-3B-Instruct} & unknown & 27/04/2025 & \href{https://huggingface.co/Qwen/Qwen2.5-7B-Instruct}{Qwen/Qwen2.5-3B-Instruct} \\
        \textsc{Qwen2.5-7B-Instruct} & unknown & 27/04/2025 & \href{https://huggingface.co/Qwen/Qwen2.5-7B-Instruct}{Qwen/Qwen2.5-7B-Instruct} \\
        \textsc{Qwen2.5-VL-32B-Instruct} & unknown & 27/04/2025 & \href{https://huggingface.co/Qwen/Qwen2.5-VL-32B-Instruct}{Qwen/Qwen2.5-VL-32B-Instruct} \\
        \textsc{Llama-3.1-8B Instruct} & 12/2023 & 23/07/2024 & \href{https://huggingface.co/meta-llama/Llama-3.1-8B-Instruct}{meta-llama/Llama-3.1-8B-Instruct} \\ 
        \textsc{Llama-3.2-3B-Instruct} & 12/2023 & 25/09/2024 & \href{https://huggingface.co/meta-llama/Llama-3.2-3B-Instruct}{meta-llama/Llama-3.2-3B-Instruct} \\   
        \textsc{Llama-3.3-70B-Instruct} & 12/2023 & 06/12/2024 & \href{https://huggingface.co/meta-llama/Llama-3.3-70B-Instruct}{meta-llama/Llama-3.3-70B-Instruct} \\            
        \hline
        \vspace{1mm}
    \end{tabular}
    \caption{\textbf{Release details of LLMs used in our experiments.} GSS (22/05/2025) and GGSS (10/01/2025) were released after the known knowledge cutoff (K-Cutoff) and even the release of the Llama-models, and less than two months before the release of the Qwen-models, making data leakage practically impossible.}
    \label{tab:model_releases}
\end{table}

\section{Appendix: Results}
\label{app:results}

\subsection{Disaggregation based on Model Family}
\label{app:results_models}

\begin{figure}
    \centering
    \subfigure[Qwen Models]{\includegraphics[width=0.9\textwidth]{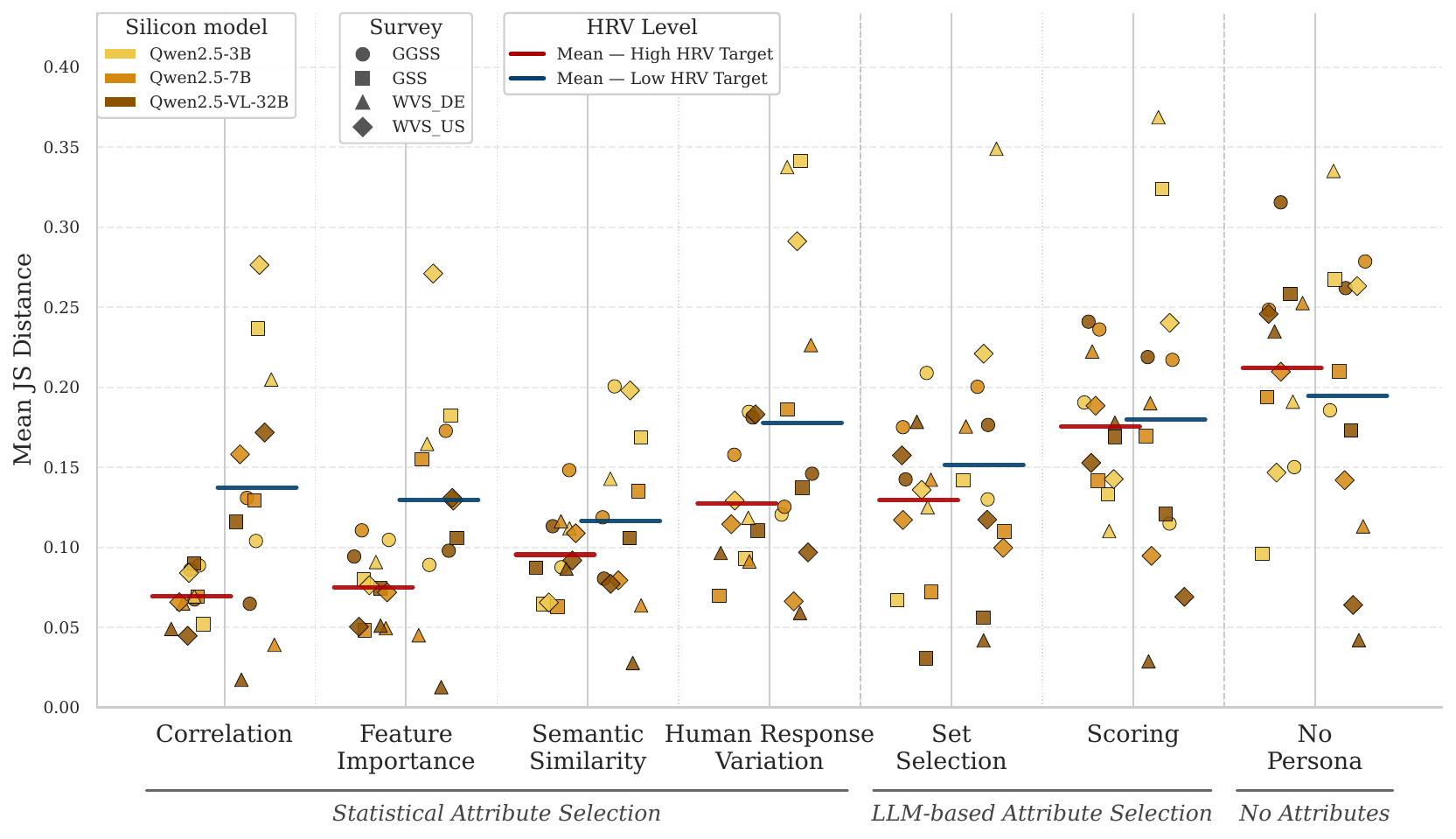}}
    \subfigure[Llama Models]{\includegraphics[width=0.9\textwidth]{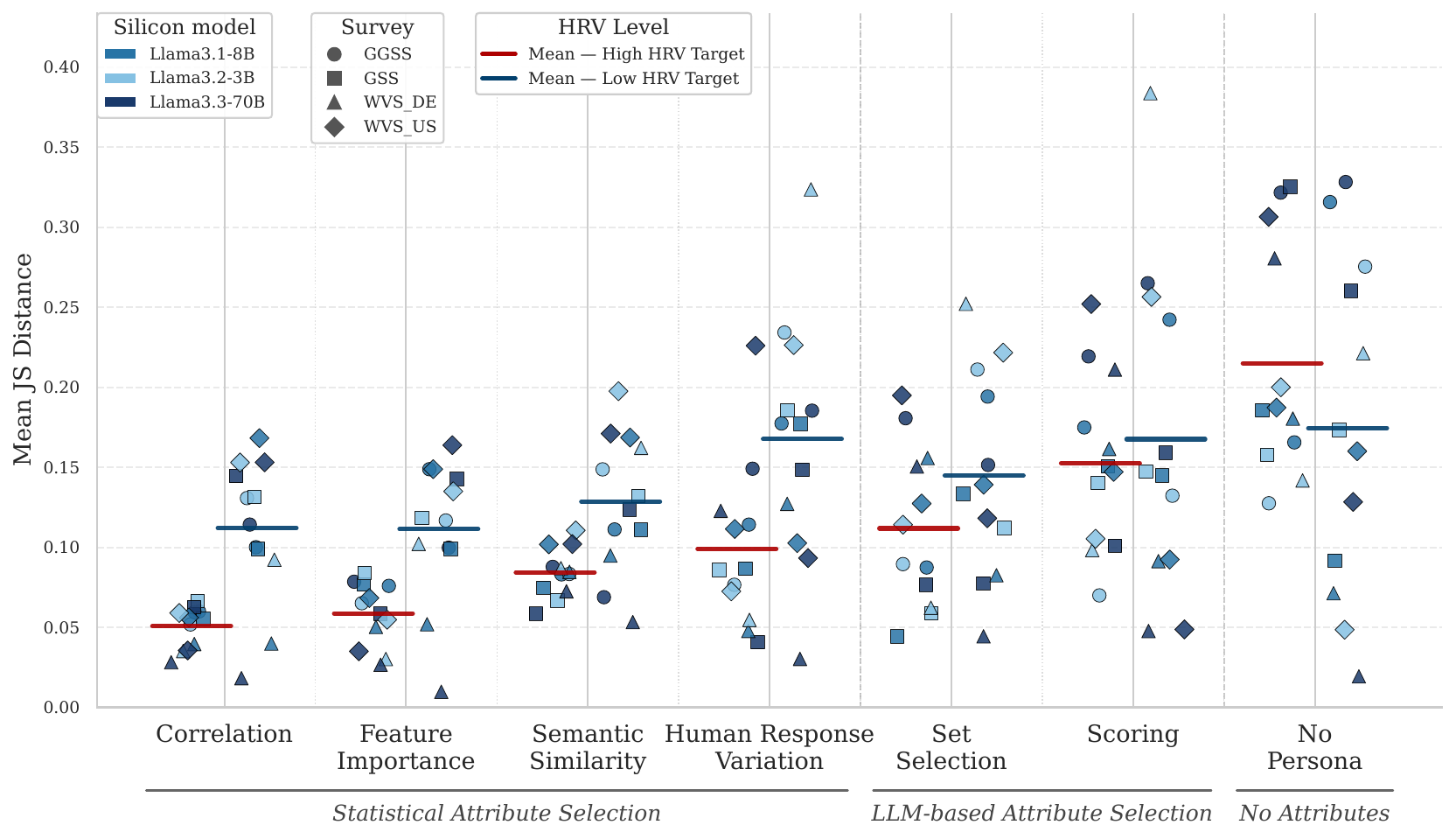}}
\caption{\textbf{Results from Figure~\ref{fig:results_overview}, disaggregated by model family.} The effects of human response variation on the performance as well as the general ordering of performance across attribute selection approaches are the same as reported for the aggregate results, implying that they are consistent across models from different families.}
\label{fig:results_models}
\end{figure}

As discussed in the main text, we find that persona-prompted LLMs are consistently better at predicting responses to target questions with high human response variation, and that, in contrast, the no persona baseline performs better for predicting target questions with low rather than high human response variation. Figure~\ref{fig:results_models} shows that this effect also holds when disaggregating based on the two model families \textemdash both Llama and Qwen models perform better in predicting responses to target questions with high human response variation than to target questions with low human response variation, and both model families show the opposite order of performance when not being conditioned on personas. 

In addition, Figure~\ref{fig:results_surveys} shows that the same effect mostly holds when disaggregating the results for the different surveys, with some exceptions. For example, the Llama models are better in predicting high response variation target questions than low response variation target questions for the GGSS, reversing the general finding. However, this seems to be survey- rather than country-specific, as the general finding still holds strongly for the German WVS data. For Qwen models, some of the attribute selection approaches (\textit{HRV}, \textit{Set Selection}, and \textit{Scoring}) perform better for low human response variation questions than high human response variation questions and thus running counter to the general finding, which again holds for the data from the other German survey, WVS\_DE. 

However, the strength of the effect, i.e., the distance between the mean lines for low and high response variation questions, is generally smaller for the surveys from Germany (GGSS and WVS\_DE) than for those from the United States (GSS and WVS\_US). Particularly when using the statistical baselines for selecting persona attributes, the ability of the LLMs to reproduce human responses for target questions with low human response variation is much worse than for target questions with high human response variation. Here, the attributes selected based on the survey data seem to move the LLMs away from the "correct" baseline response, compared to attributes selected through other methods.

\subsection{Disaggregation based on Survey}
\label{app:results_surveys}

\begin{figure}
    \centering
    \subfigure[Qwen Models]{\includegraphics[width=0.85\linewidth]{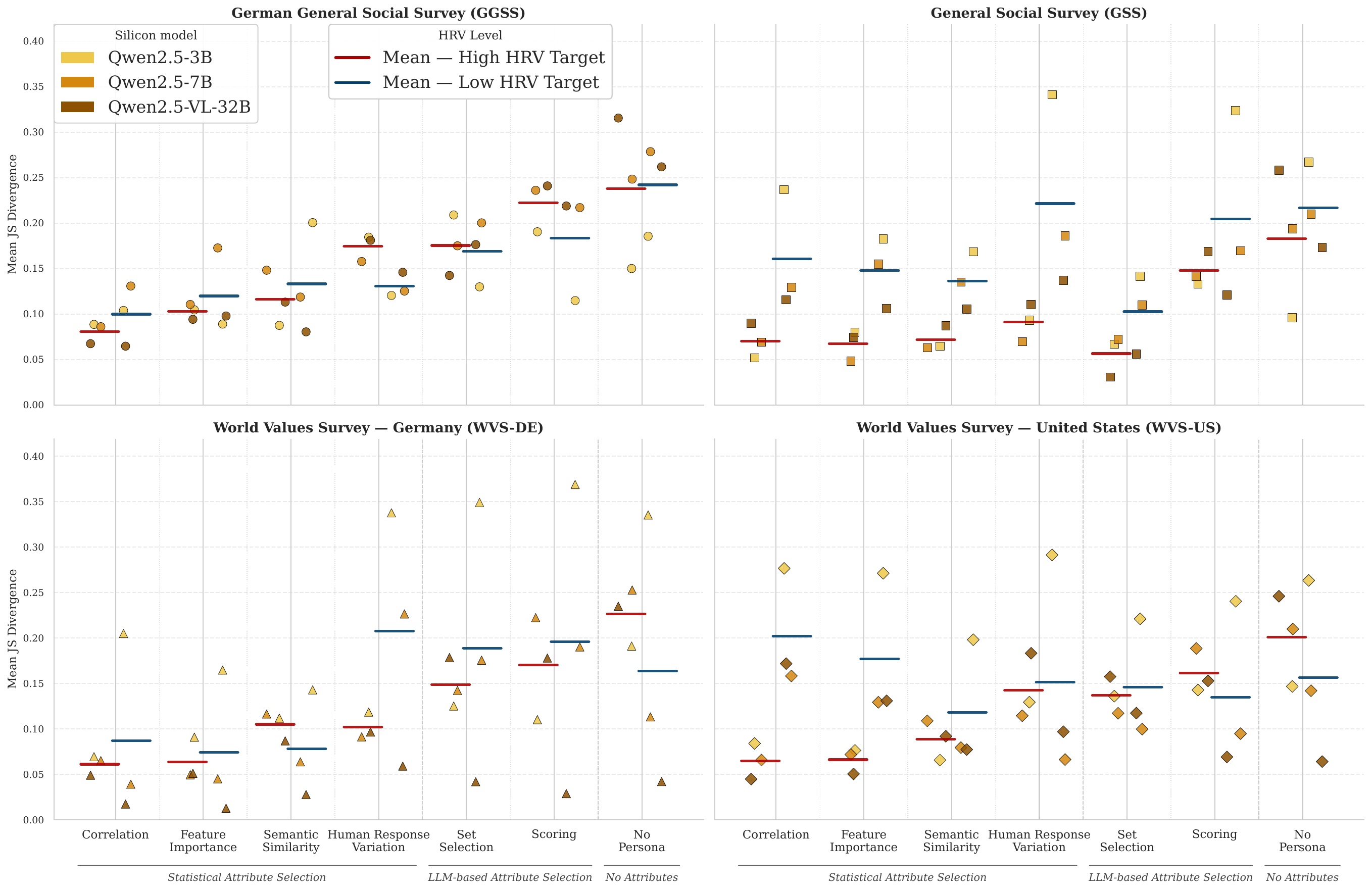}}
    \subfigure[Llama Models]{\includegraphics[width=0.85\linewidth]{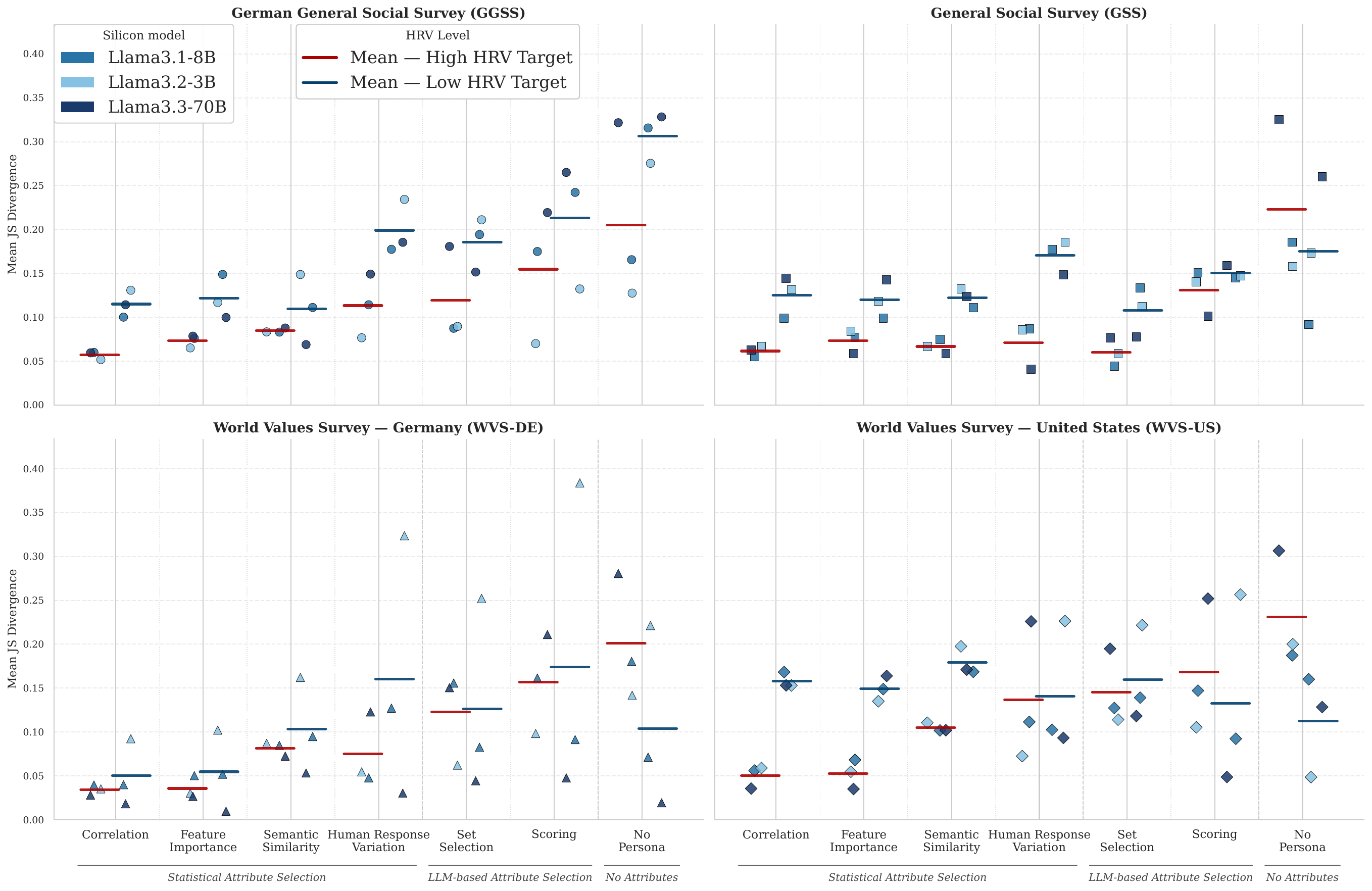}}
    \caption{\textbf{Results from Figure~\ref{fig:results_overview}, disaggregated by model family and survey.} With some exceptions, e.g., the reversal of the effect of human response variation in the no persona baseline for GGSS data in Llama models and for some attribute selection methods in Qwen models, the general findings reported for the aggregated results hold true also across surveys.}
    \label{fig:results_surveys}
\end{figure}

We already discuss in the main text that the statistical baselines for selecting attributes perform better than the LLM-based selection approaches. Figure~\ref{fig:results_models} shows that the same performance order of persona attribute selection methods persists when disaggregating the results based on model families, with performance differences being particularly large for high response variation target questions. Figure~\ref{fig:results_surveys} shows that the performance order of attribute selection approaches is also consistent across the different surveys, with differences between attribute selection approaches generally being less pronounced for the US-based surveys (GSS and WVS\_US).

\newpage

\subsection{Stability of LLM-based Attribute Selection Approaches}
\label{app:results_stability}

\subsubsection{Stability of the Set Selection Approach} 
\label{app:results_stability_setselection}

\paragraph{Evaluation Approach}
For the LLM-based set selection approach, we count how often each survey variable is selected among the five most important attributes for predicting a given target question across the 100 (30) independent runs. We order these selection counts in decreasing order and plot them per LLM for all 20 target questions across the four surveys, with the x-axis corresponding to the variables in a survey and the y-axis to the selection counts. For a perfectly stable selection method, we would expect every run to select the same set of five attributes as most important for each target question, and for a selection method that is random guessing, we would expect every variable to be selected equally often. We can therefore visually assess and compare the stability of the set selection approach across different LLMs.

\paragraph{Results}

\begin{figure}[h]
    \centering
    \subfigure[Llama-3.2-3B-Instruct]{\includegraphics[width=60mm]{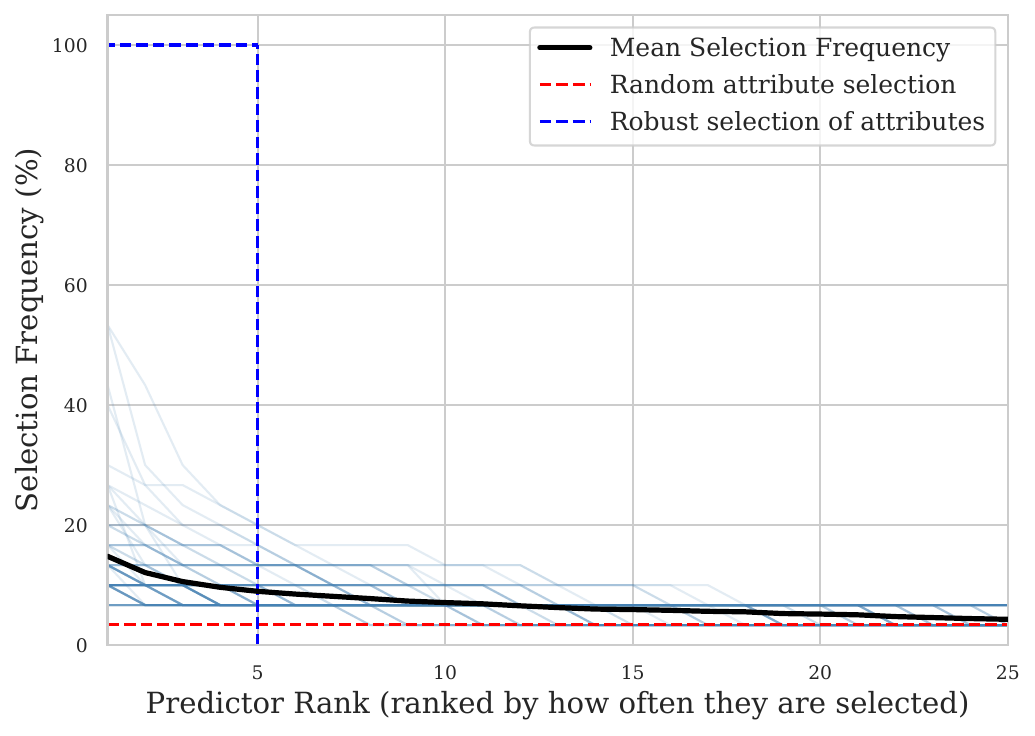}}
    \hfill
    \subfigure[Qwen2.5-3B-Instruct]{\includegraphics[width=60mm]{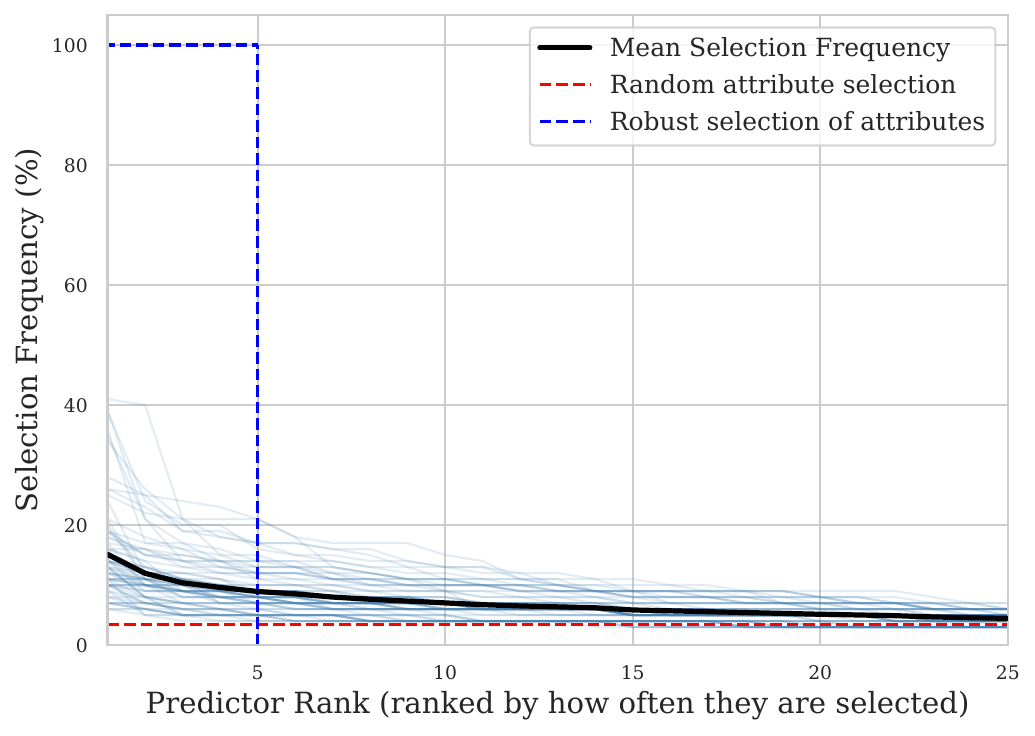}}

    \subfigure[Llama-3.1-8B-Instruct]{\includegraphics[width=60mm]{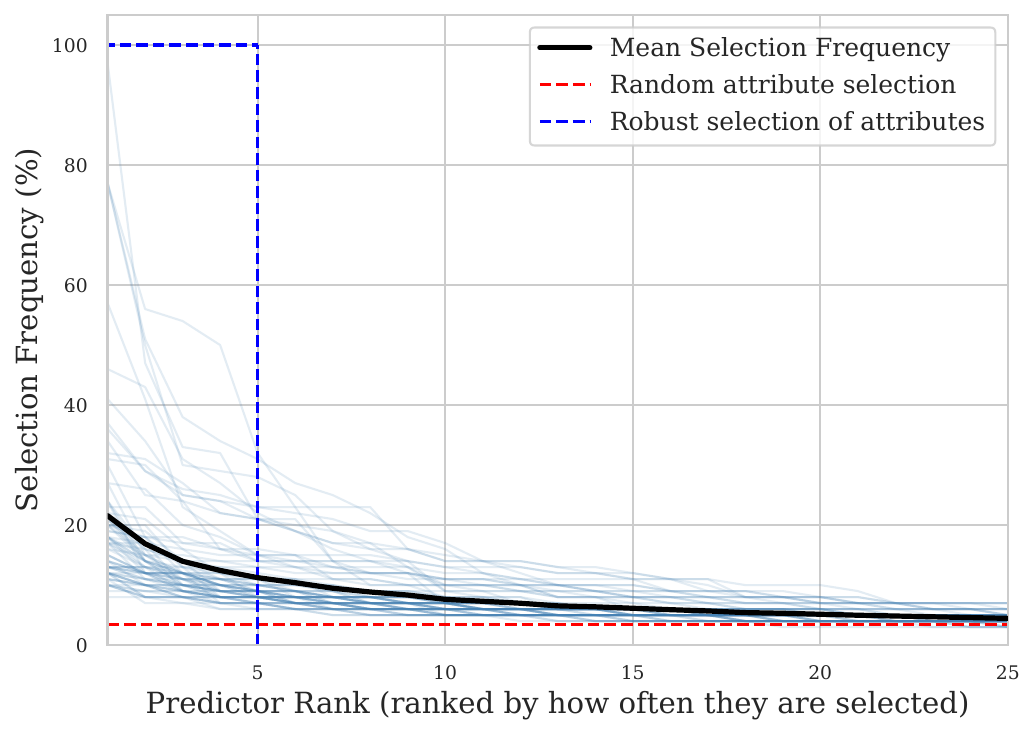}}
    \hfill
    \subfigure[Qwen2.5-7B-Instruct]{\includegraphics[width=60mm]{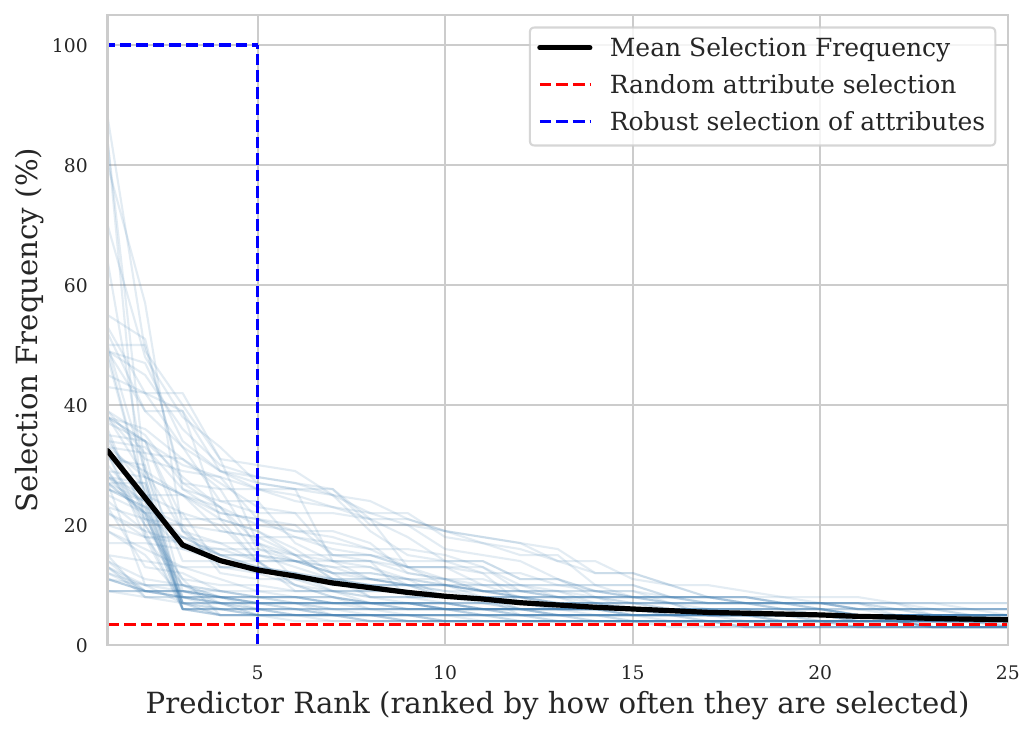}}

    \subfigure[Llama-3.3-70B-Instruct]{\includegraphics[width=60mm]{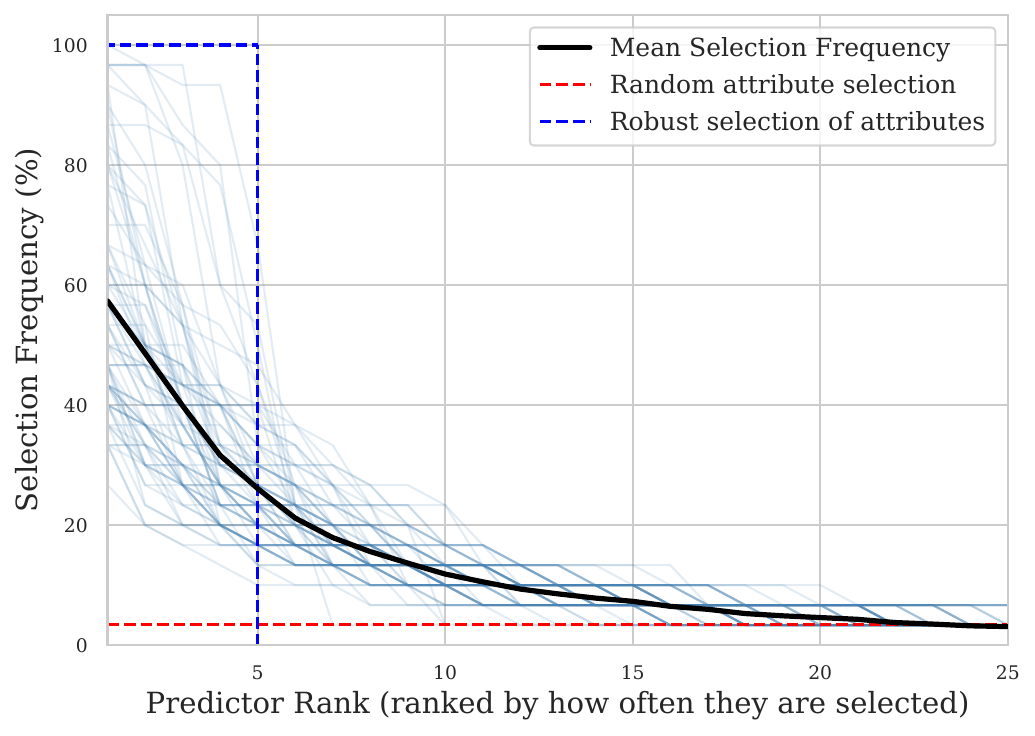}}
    \hfill
    \subfigure[Qwen2.5-VL-32B-Instruct]{\includegraphics[width=60mm]{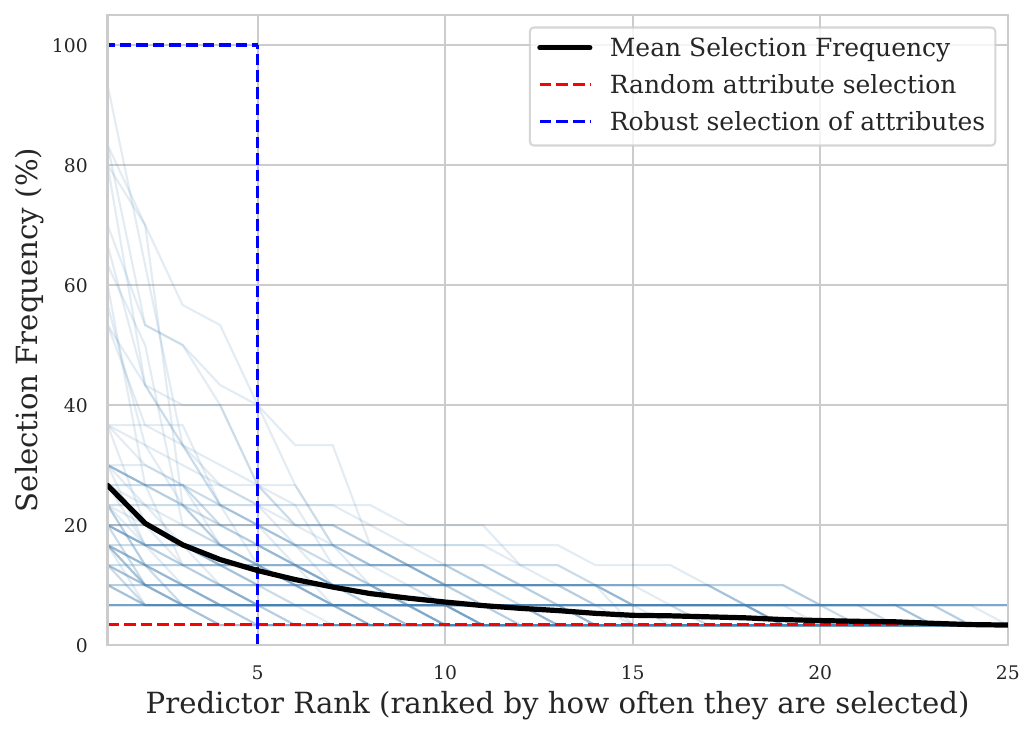}}

	\caption{\textbf{Results from Figure~\ref{fig:selection_robustness_mean}, disaggregated by model.} The increased stability of models with higher parameter counts clearly shows, and particularly the ability of larger models to select the same attribute as most important for some target questions in more than 80\% of runs.}
	\label{fig:results_stability_setselection}
\end{figure}

As discussed in the main text, the stability of the set selection method increases with model size. For the smallest models in each family, the same attribute is on average selected as most important in only 15.1\% (Qwen2.5-3B) and 14.8\% (Llama3.2-3B) of runs conducted for each target questions, indicating a large variety of attributes considered as important across the runs. 

For the mid-sized models, the stability increases substantially\textemdash Qwen2.5-7B now selects the same attribute as most important in 32.4\% and Llama3.1-8B in 21.6\% of the runs per target questions. Additionally, we see that for these models some target questions achieve very high consistency across runs, with the most important attribute for some few target questions being the same in nearly every run.

For Llama3.3-70B we see high stability of the attribute selection across runs. Not only does it consistently select the same attribute as most important in on average 57.3\% of runs across target questions, but it even selects all five attributes consistently for a number of target questions. For Qwen2.5-32B, the stability of the set selection is similar to that of the mid-sized version of the model, with the most important attribute being selecting in roughly a third of the runs across target questions.

\FloatBarrier
\clearpage

\subsubsection{Stability of the Scoring Approach} 
\label{app:results_stability_scoring}

\paragraph{Evaluation Approach}
For the LLM-based scoring approach, we create a ranking of the variable importance scores for each of the independent 100 (30) runs. From these rankings, we then construct a distribution of concordance coefficients for each LLM scoring the 20 target questions across the four surveys. We compare the means and variances of these distributions to get a sense of the stability of the scoring approach across different LLMs. As we have to account for the possibility of predictors with the same score and thus sharing the same rank, we calculate the tie-corrected Kendall's $W$ for each of the 100 (30) runs.

\paragraph{Results}

\begin{figure}
    \centering
    \subfigure[Llama-3.2-3B-Instruct]{\includegraphics[width=0.8\linewidth]{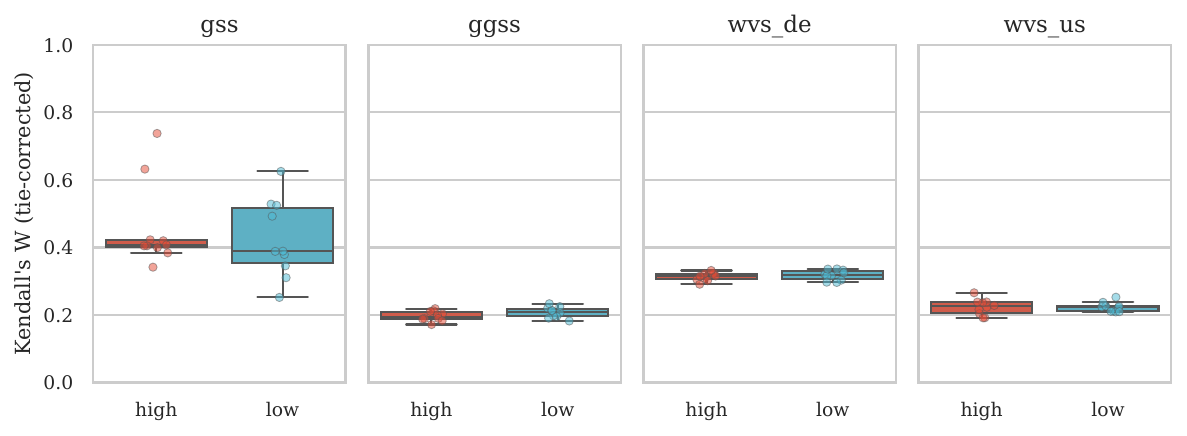}}
    \subfigure[Llama-3.1-8B-Instruct]{\includegraphics[width=0.8\linewidth]{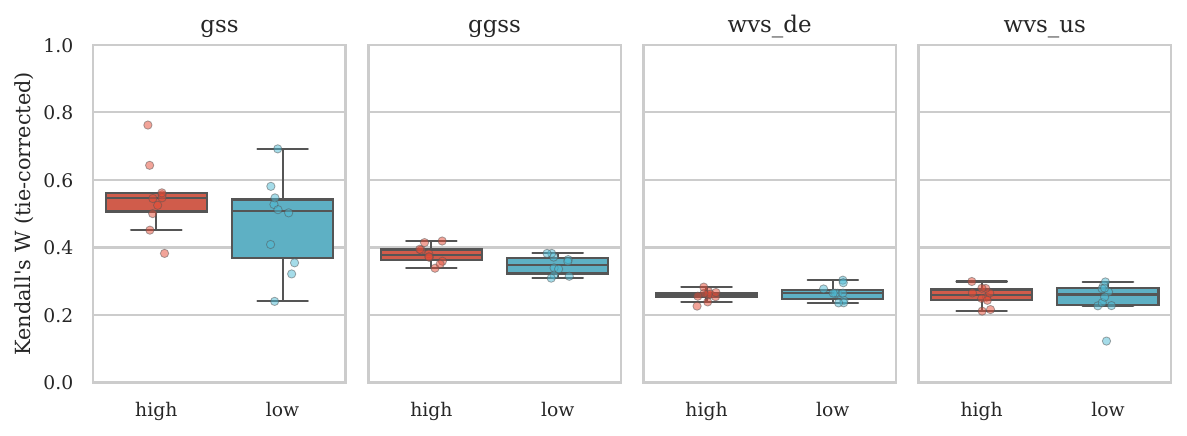}}

    \subfigure[Llama-3.3-70B-Instruct]{\includegraphics[width=0.8\linewidth]{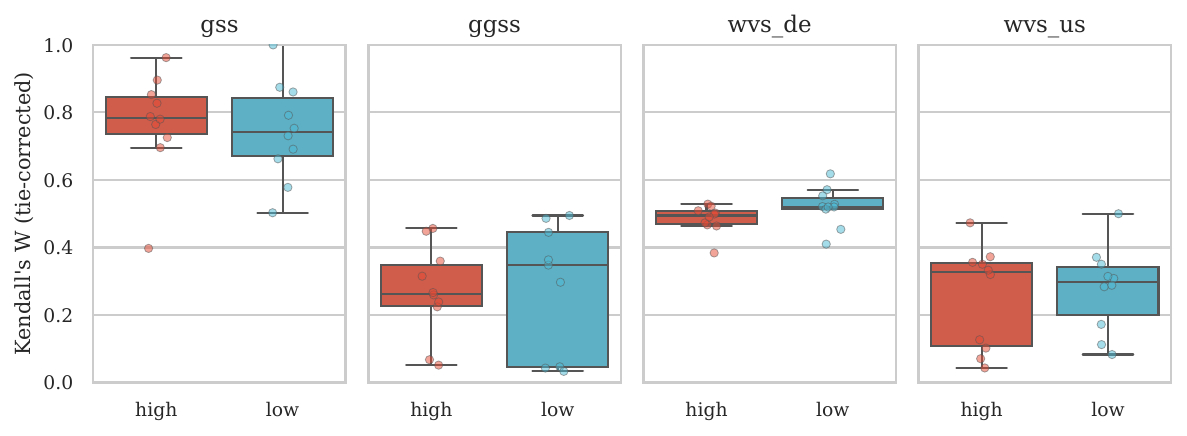}}
	\caption{\textbf{Results from Figure~\ref{fig:scoring_robustness_mean}, disaggregated by model family (Llama).}}
	\label{fig:results_stability_scoring_llama}
\end{figure}

For models from the Llama family, Figure~\ref{fig:results_stability_scoring_llama} shows that the concordance when scoring the importance of survey items from the GSS is much higher compared to all other surveys. This difference can potentially be explained through the fact that the pool of predictors and thus the number of possible rank variations is smallest for the GSS. This makes it theoretically easier to achieve higher Kendall's $W$ values. Surprisingly, the largest Llama model seems to be the most erratic model across different surveys, with scoring predictors very consistently across multiple runs for the GSS and showing very high variation but generally low consistency for GGSS and WVS\_DE.

\begin{figure}
    \centering
    \subfigure[Qwen2.5-3B-Instruct]{\includegraphics[width=0.8\linewidth]{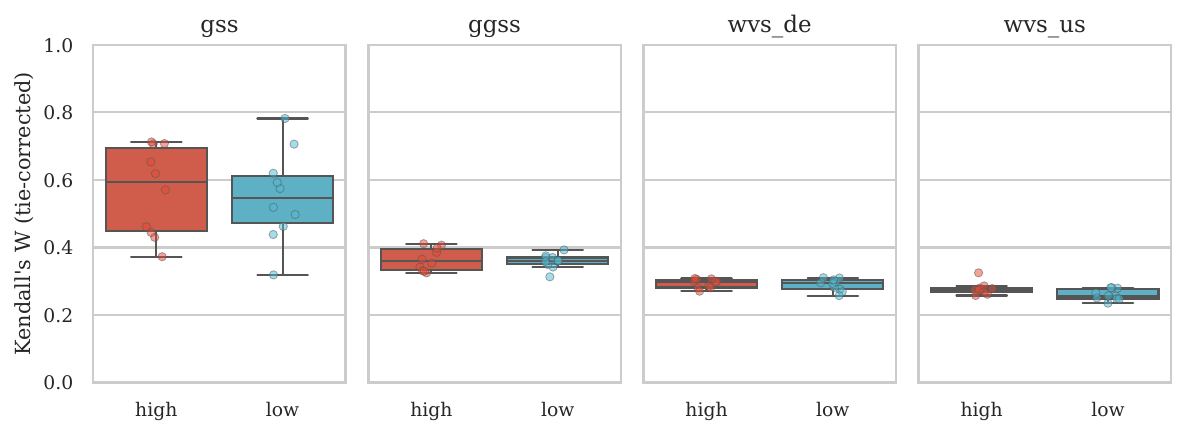}}

    \subfigure[Qwen2.5-7B-Instruct]{\includegraphics[width=0.8\linewidth]{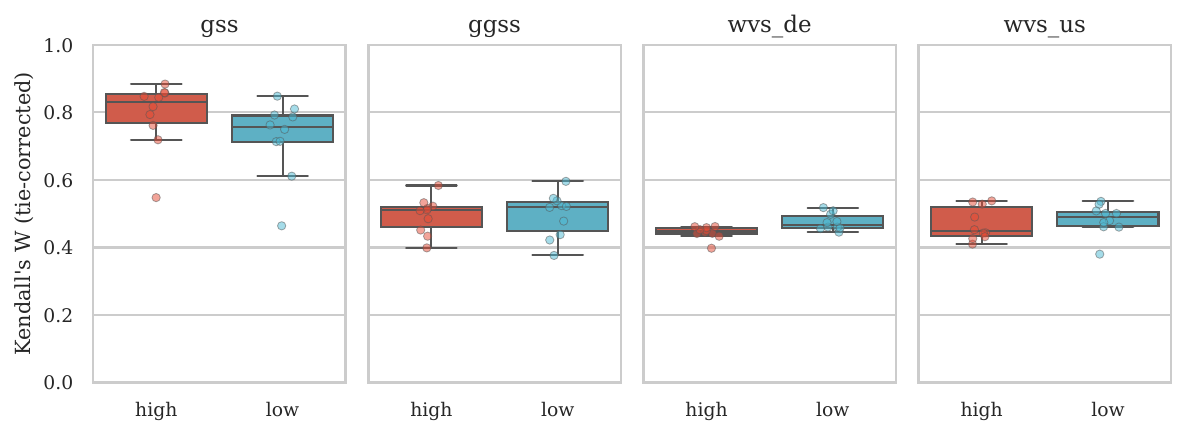}}
    \subfigure[Qwen2.5-VL-32B-Instruct]{\includegraphics[width=0.8\linewidth]{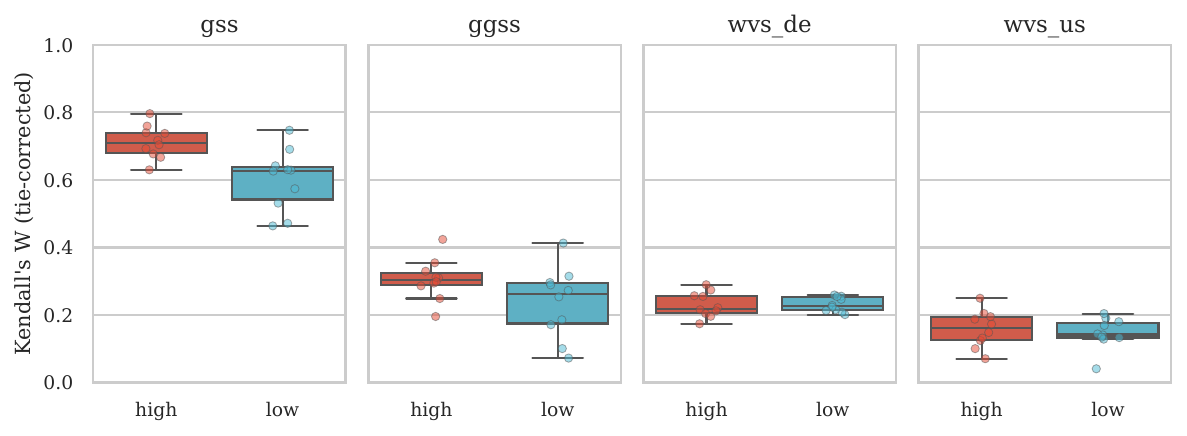}}

	\caption{\textbf{Results from Figure~\ref{fig:scoring_robustness_mean}, disaggregated by model family (Qwen).}}
	\label{fig:results_stability_scoring_qwen}
\end{figure}

For models from the Qwen family, Figure~\ref{fig:results_stability_scoring_qwen} shows that there are barely any survey-specific differences between the surveys GGSS and WVS (USA \& GER). However, the concordance for all models in the Qwen family when scoring items from the General Social Survey is much higher compared to all other surveys. This could again be due to the fact that GSS has the lowest number of survey variables and thus the lowest number of possible rankings. Differences between models sizes are less pronounced for the Qwen models than for the Llama models.

\end{document}